\documentclass{article}

\usepackage{fullpage} 
\usepackage[separate-uncertainty=true,multi-part-units=single]{siunitx} 
\usepackage{appendix} 

\usepackage{cite}
\usepackage{amsmath,amssymb,amsfonts}

\usepackage{multirow}
\usepackage{makecell}
\usepackage{booktabs}
\usepackage{tabulary}

\usepackage{graphicx}
\usepackage{subcaption}
\usepackage{textcomp}
\usepackage[dvipsnames]{xcolor}         
\usepackage{algpseudocode}
\usepackage{algorithm}
\usepackage{algorithmicx}
\usepackage{url}

\usepackage{hyperref}

\usepackage{soul}
\def\BibTeX{{\rm B\kern-.05em{\sc i\kern-.025em b}\kern-.08em
    T\kern-.1667em\lower.7ex\hbox{E}\kern-.125emX}}
\usepackage{bm}
\usepackage{enumitem}

\newcommand{\Reviewer}{r}
\newcommand{\Paper}{p}

\newcommand{\Title}{t_{p}}
\newcommand{\Abstract}{a_{p}}
\newcommand{\AbstractAdv}{a^{\text{adv}}_{p}}
\newcommand{\PaperAbstractAdv}[1]{a^{\text{adv}}_{#1}}
\newcommand{\AbstractAdvEdit}{{\PaperAbstractAdv{\Paper}}'}
\newcommand{\AbstractAdvVersion}[1]{{\PaperAbstractAdv{\Paper}}^{(#1)}}

\newcommand{\PastPaper}{q}

\newcommand{\Archive}[1]{Q_{\Reviewer}}
\newcommand{\ArchiveAdv}[1]{\Archive{}^{\text{adv}}}

\newcommand{\Similarity}[2]{s({#1},{#2})}

\newcommand{\SimilarityAbstract}[3]{s((#1,#2),#3)}
\newcommand{\PaperReviewerPair}{(\Paper,\Reviewer)}
\newcommand{\HumanMode}{\textit{human-in-the-loop}}
\newcommand{\AutoMode}{\textit{fully automatic}}

\newcommand{\Embedding}[1]{\mathbf{v}_{#1}}
\newcommand{\norm}[1]{\left\lVert#1\right\rVert_2}
\newcommand{\Regenerations}{\bm{N}}
\newcommand{\Alternations}{{\bm M}}
\newcommand{\Keywords}{{\bm K}}
\newcommand{\AllPapers}{\mathcal{P}}
\newcommand{\AllReviewers}{\mathcal{R}}

\newcommand{\IncludeRelatedWorkFunctionName}{\textsc{\textbf{IncludeThemes}}}
\newcommand{\IncludeRelatedWork}{{\IncludeRelatedWorkFunctionName{}}}
\newcommand{\InsertKeywordsFunctionName}{\textsc{\textbf{InsertKeywords}}}

\newcommand{\InsertKeywords}{{\InsertKeywordsFunctionName{}}}

\newcommand{\FindKeywords}{\textsc{FindKeywords}}

\newcommand{\ConstraintsCheck}[1]{\textbf{ConstraintCheck}(#1)}
\newcommand{\SimilarityCheck}[5]{\textbf{SimilarityCheck}(#1,#2,#3,#4,#5)}
\newcommand{\EarlyStoppingCheck}[2]{\textbf{EarlyStoppingCheck}(#1,#2)}

\newcommand{\SimDelta}{\delta}
\newcommand{\KeywordsList}{\textit{keywords}}

\newcommand{\word}[1]{w_{#1}}

\newcommand{\ArchiveWords}{W}

\newcommand{\ithMaxSimWord}[1]{w_{\max}^{(#1)}}

\DeclareMathOperator*{\argmax}{arg\,max}
\algnewcommand{\IIf}[1]{\State\algorithmicif\ #1\ \algorithmicthen}
\algnewcommand{\EndIIf}{\unskip\ \algorithmicend\ \algorithmicif}

\newcommand{\floatfootnote}[1]{\ifx\[$\else\footnote{#1}\fi}
    
\begin{document}

\title{Vulnerability of Text-Matching in ML/AI\\Conference Reviewer Assignments to Collusions
}

\author{
Jhih-Yi (Janet) Hsieh, Aditi Raghunathan, Nihar B. Shah\\{\small School of Computer Science, Carnegie Mellon University}\\ {\tt \small jhihyih@alumni.cmu.edu, \{aditirag, nihars\}@andrew.cmu.edu}
}

\date{}

\maketitle

\begin{abstract}
In the peer review process of top-tier machine learning (ML) and artificial intelligence (AI) conferences, reviewers are assigned to papers through automated methods. These assignment algorithms consider two main factors: (1) reviewers' expressed interests indicated by their bids for papers, and (2) reviewers' domain expertise inferred from the similarity between the text of their previously published papers and the submitted manuscripts. A significant challenge these conferences face is the existence of collusion rings, where groups of researchers manipulate the assignment process to review each other's papers, providing positive evaluations regardless of their actual quality. Most efforts to combat collusion rings have focused on preventing bid manipulation, under the assumption that the text similarity component is secure. In this paper, we demonstrate that even in the absence of bidding, colluding reviewers and authors can exploit the machine learning based text-matching component of reviewer assignment used at top ML/AI venues to get assigned their target paper. We also highlight specific vulnerabilities within this system and offer suggestions to enhance its robustness.
\end{abstract}

\section{Introduction}
The rapid growth of machine learning and artificial intelligence research has led to major scientific conferences in these fields receiving thousands to tens of thousands of paper submissions and similar numbers of reviewers~\cite{shah2022challenges}. To manage this overwhelming volume, the assignment of reviewers to papers has become largely automated. A key component of this automated assignment pertains to text-matched similarity scores between reviewers' past work and submitted manuscripts, where natural language processing algorithms compute the similarity between the text of each submitted manuscript and the texts of the reviewer's previously published manuscripts. Models like SPECTER~\cite{cohan2020specter}, which generates embeddings based on textual content, have become widely used across various prestigious venues, including the 
Conference on Neural Information Processing Systems (NeurIPS), and are a default model on the popular OpenReview conference management platform.\footnote{While our work focuses on the peer-review process in ML/AI venues, the continuous evolution and scaling up of other communities suggest that they may soon have to adopt similar processes and face similar challenges.}

Although automation facilitates efficiency, it also introduces potential vulnerabilities. One major challenge threatening the integrity of the peer review process is the existence of \emph{collusion rings} -- groups of individuals who conspire to manipulate the review system for personal gain~\cite{littman2021collusion, vijaykumar2020ieee}. These rings can unfairly influence the acceptance of certain papers by orchestrating favorable reviews from colluding reviewers.
As noted in \cite{vijaykumar2020ieee}, colluders may attempt to ``game the system by exploiting vulnerabilities in the assignment algorithms to have their collaborators review their submissions.'' 

Studies to identify or mitigate collusion have thus far focused on analyzing manipulations in \emph{bidding}, the part of the reviewer assignment process where reviewers can indicate their interest in reviewing each paper~\cite{wu2021making,jecmen2024detection}. The algorithm in \cite{wu2021making} considers text similarities as ground truth when examining the bids. In practice, program chairs have also focused on bidding when investigating collusion rings \cite{ICLR2024chairs}; to address concerns over bidding manipulation, venues such as the Conference on Computer Vision and Pattern Recognition (CVPR) and ACL Rolling Review (ARR) have removed the bidding process altogether. This focus on bidding implicitly or explicitly assumes that the text-matching algorithms resistant to manipulation. However, this assumption warrants scrutiny. 

In this paper, we investigate whether the text-matching algorithms used in automated reviewer assignments are indeed robust against manipulation.  
Since most papers are assigned to 3--6 reviewers at conferences \cite{shah2022challenges}, we focus our evaluation on whether a pair of colluding author and reviewer can successfully manipulate the text-matching algorithms to give the colluding reviewer one of the top-1, top-3, or top-5 highest text similarity to the colluding paper among all reviewers at a conference. We find that the text matching algorithms are susceptible to attacks that can significantly increase the calculated similarity between a paper and its colluding reviewer. Here are some salient details:

\begin{itemize}
    \item We find that the SPECTER \cite{cohan2020specter} text similarity matching algorithm, used by top ML/AI conferences for reviewer assignments, can be manipulated by colluding authors and reviewers. 
    In evaluations on NeurIPS 2023 data, our attack successfully increases a colluding reviewer's similarity from ranked 101 to top-5 for a colluding paper about $92\%$ of the time.
    \item Considering NeurIPS 2022 as a ``past'' conference whose data is publicly available to an attacker, and NeurIPS 2023 as the ``current'' conference whose data is unavailable, we find that the attack performance on NeurIPS 2022 is reflective of the performance in NeurIPS 2023. We find strong correlations of 0.62, 0.83, 0.92 and 0.93 between colluding reviewers' similarity rankings in NeurIPS 2022 and NeurIPS 2023 in four different settings. 
    \item Many conferences allow reviewers to select a subset of papers in their profile. We find that a careful selection of the past papers by the reviewer -- and particularly a choice of fewer papers -- can significantly increase the attack's success rate. When a colluding reviewer selects only one paper that is the most similar to the colluding paper, it increases the colluding reviewer's similarity from ranked 101 to top-5 for the colluding paper 41\% of the time even without any modifications to the colluding paper's abstract.
    \item When reviewers have multiple papers in their profile, similarity-computation algorithms which assign papers via the maximum of its similarities with the reviewer's past papers are more vulnerable than those which take the mean of the similarities with the reviewer's past papers. When the mean of similarities is taken, the attack successfully increases a colluding reviewer's similarity from ranked 101 to top-5 for a colluding paper 32\% of the time compared to 49\% when the maximum is taken.
    \item We conduct a human-subject experiment to test for the identifiability of adversarial abstracts, in which researchers are asked to evaluate manipulated or control abstracts. For scalability, this experiment uses only automated attacks without the human-in-the-loop interventions, thereby obtaining an upper bound on the detectability (as attackers would otherwise check and iterate if they think the attack is detectable). We find that the rate at which participants complain about coherence or consistency of the abstracts is higher in the attacked abstracts, but there is also a non-trivial rate of complaints about the control abstracts, which in practice can give attackers plausible deniability.
\end{itemize}
In the final section of the paper,  we discuss the implications of our findings and propose recommendations to enhance the security and robustness of automated reviewer assignments systems. We have also informed the concerned conference management platform about these results. Based on the findings of this work, the platform has implemented safeguards for conference organizers to guard against such attacks, and the safeguards have been used by a major AI/ML conference. Subsequent peer-review venues have also expressed interest in deploying safeguards, and we are closely helping these venues in doing so.

We release our research artifacts including code, datasets, and manipulation examples on Zenodo \url{https://doi.org/10.5281/zenodo.15588237}. This paper aims to make the community aware of this vulnerability, which has implications for the fairness and integrity of the peer review process.

\section{Problem setting}

In this section, we describe the automated reviewer assignments that are widely used in ML/AI conferences, along with our problem setting and the attacker's threat model.

\subsection{Automated Reviewer Assignments}\label{sec:assignments}

We consider peer review at conferences, which are the primary venue of publication in computer science. In conference review, there is a pre-chosen pool of reviewers, and we denote this set of all reviewers as $\AllReviewers$. We let $\AllPapers$ denote the set of all submitted papers. To assign reviewers to papers, the program chairs compute a \textbf{``similarity score''} for each reviewer-paper pair. The similarity score is a number between 0 and 1, and a higher value of the similarity score indicates a higher envisaged expertise of that reviewer for that paper. 

The similarity score is often a combination of multiple components. A primary component is text matching between the submitted paper and the reviewer's previous papers, which is the focus of this paper and is discussed below. Another important component is bidding, where each reviewer can see the list of submitted papers and indicate their interest in reviewing each paper. 
Prior studies~\cite{jecmen2020mitigating,wu2021making} have demonstrated that bidding is easily susceptible to manipulation.  
Most previous research on guarding against collusion rings~\cite{jecmen2020mitigating,wu2021making,jecmen2022tradeoffs,jecmen2022dataset, jecmen2024detection, leyton2024matching}, as well as practical implementations (e.g., \cite{ICLR2024chairs, leyton2024matching}), have focused on bidding. 
For example, venues like CVPR and ARR have completely disabled bidding to prevent manipulation. 
Furthermore, {\bf it is generally assumed that bidding can be gamed, while text similarity-based methods are considered robust} (e.g., in \cite{wu2021making}, as well as by venues that have banned bidding in favor of text similarity approaches). A third component involves the subject areas or keywords provided by the authors and reviewers, though these can also be gamed~\cite{ailamaki2019sigmod}.

Reviewers are then assigned to papers by solving an optimization problem to choose the assignment with the highest similarity scores, subject to various constraints, such as reviewer and paper loads and known conflicts of interest. 
A higher similarity score for any reviewer-paper pair means this reviewer is more likely to get assigned the paper. 
To answer if reviewer assignments can be manipulated for a colluding paper-reviewer pair, we evaluate the increase in the colluding reviewer's similarity score ranking when compared to other reviewers for the paper. Since different conferences have different optimization programs~\cite{charlin2013toronto, stelmakh2018forall, kobren19localfairness, jecmen2020mitigating, payan2021will, leyton2024matching}, evaluating the similarity rankings, which are relative, makes our results more generally applicable. With this context, we focus on the vulnerability of the similarity score that depends only on text similarity, since bidding is known to be vulnerable and can only increase the success of reviewer assignment manipulations. 

\subsection{Paper-Reviewer Text Similarity} 
\label{sec:text_sim}
For a $\Paper \in \AllPapers$ and reviewer $\Reviewer \in \AllReviewers$, we let $\Similarity{\Paper}{\Reviewer} \in [0,1]$ denote the text similarity between this reviewer and paper. This similarity can be computed in many ways~\cite{charlin2013toronto,cohan2020specter, chang2021mfr, singh2022scirepeval, ostendorff2022neighborhood}. In this work, we focus on the widely used SPECTER algorithm~\cite{cohan2020specter}.

SPECTER is widely used for assigning reviewers in ML/AI related venues such as NeurIPS, the International Conference on Machine Learning (ICML), the International Conference on Learning Representations (ICLR) and the Transactions on Machine Learning Research (TMLR) journal. SPECTER uses neural network embeddings trained with an objective to keep the embeddings of similar papers close, where two papers are considered similar when one of them cites the other. 

For any reviewer $r \in \AllReviewers$, we let $\Archive{\Reviewer}$ denote as the default archive containing $10$ most recent papers authored by $\Reviewer$. The similarity score $\Similarity{\Paper}{\Reviewer}$ is an aggregation of the cosine similarities between the SPECTER embedding of $\Paper$ and the SPECTER embeddings for each of the different papers in $\Archive{\Reviewer}$. Formally, for any paper $\Paper$, let $\Embedding{\Paper} \in \mathbb{R}^{768}$ denote its SPECTER embedding vector. In practice, only the titles and abstracts are used to compute $\Embedding{\Paper}$. Then, for a submitted paper $\Paper \in  \AllPapers$ and each paper $\PastPaper \in \Archive{\Reviewer}$ in a reviewer's archive, the individual similarity between $\Paper$ and $\PastPaper$ is computed as the cosine similarity between their embeddings: $\frac{\Embedding{\Paper} \cdot \Embedding{\PastPaper}}{\norm{\Embedding{\Paper}} \norm{\Embedding{\PastPaper}}}$.

The protocol for handling multiple papers in the reviewer's archive $\Archive{\Reviewer}$ involves the following two steps: (i) A similarity score is computed between the submitted paper and each individual paper in the reviewer's archive (described above); and (ii) To determine an overall similarity score between the submitted paper and the reviewer, the individual similarity scores across the papers in the reviewer's archive are then aggregated. Two commonly used aggregation methods are:
\begin{itemize}
    \item Average (mean) pooling: $\Similarity{\Paper}{\Reviewer} = \frac{1}{\lvert \Archive{\Reviewer} \rvert} \underset{\PastPaper \in \Archive{\Reviewer}}{\sum} \frac{\Embedding{\Paper} \cdot \Embedding{\PastPaper}}{\norm{\Embedding{\Paper}} \norm{\Embedding{\PastPaper}}}$, and
    \item Max pooling: $\Similarity{\Paper}{\Reviewer} = \underset{\PastPaper \in \Archive{\Reviewer}}{\max} \frac{\Embedding{\Paper} \cdot \Embedding{\PastPaper}}{\norm{\Embedding{\Paper}} \norm{\Embedding{\PastPaper}}}$.
\end{itemize}

Max pooling and average pooling are both commonly used, with max pooling generally preferred because it has been observed to provide more accurate similarity scores.

In addition, some conferences may have $\Archive{\Reviewer}$ containing a different number of most recent papers by default. In fact, it is common to allow reviewers to \emph{curate their own archives}, which as we demonstrate provides an attack surface.

\subsection{Attacker's Threat Model} \label{sec:threat_model}
In this section, we describe a realistic threat model that directly applies to standard conferences that use automated reviewer assignments based on SPECTER. 

\paragraph{\bf Attacker objective} 
The ``attacker'' represents both the colluding author of a paper $\Paper \in \AllPapers$ and a reviewer $\Reviewer \in \AllReviewers$ working together to ensure $\Paper$ is assigned to $\Reviewer$ for peer review. For text-similarity-based assignments, this corresponds to increasing $\Similarity{\Paper}{\Reviewer}$ so that $\Reviewer$ ranks in top-k of the conference's reviewers ranked by their similarity scores with $\Paper$. In our subsequent experiments, we consider $k \in \{1, 3, 5\}$. 

\paragraph{\bf Attack surface and constraints} We assume that each pair of colluding author and reviewer (the ``attacker'') know each other and are actively colluding, so they can work together to create the attack. The colluding author can manipulate the abstract of their paper, and the colluding reviewer can adversarially curate their reviewer archive $\ArchiveAdv{\Reviewer}$ to only retain selected papers. We only consider abstract modifications, since the 
SPECTER similarity scores are computed based on title and abstract only, though the author can change any parts of their paper. 
In this paper, we also do not consider changes to the title because we suspect title modifications may arouse more suspicion from non-colluding reviewers.

When modifying the abstract, the attacker must avoid arousing suspicion in non-colluding reviewers. This is ostensibly vague and we formalize this notion via two constraints on adversarial modifications of the abstract: (i) \textit{Coherence}: The abstract should use academic writing style, have natural flow, and cannot contain scientifically false information. (ii) \textit{Consistency}: The abstract should be consistent with the paper's contents. 
In this paper, we also enforce constraints on the number of sentences and keywords that can be added to the colluding paper's abstract. 
Attacks are generally allowed only one sentence that is related to $\ArchiveAdv{\Reviewer}$, but we allow up to three sentences related to $\ArchiveAdv{\Reviewer}$ in some cases only when there is manual supervision involved to ensure coherence and consistency. In addition, attacks are allowed to add up to 10 keywords selected from $\ArchiveAdv{\Reviewer}$.  
We also perform human-study experiments to judge whether non-colluding reviewers find that abstracts generated by our proposed attack are suspicious. 

\paragraph{\bf Attacker access} The attackers have access to the exact SPECTER embeddings, with the model weights publicly available at \url{https://github.com/allenai/specter}. For many conference venues, the attackers also have access to the publicly available reviewer pool from the previous year's conference (e.g.,  \url{https://neurips.cc/Conferences/2023/ProgramCommittee}). However, they do not have access to the reviewer pool of the current iteration of the conference.

\section{Related work}
\label{SecRelatedWork}

We now discuss literature that is closely related to our work.

\paragraph{\bf Text matching for automated assignment} There are a number of algorithms for computing text matching similarity scores~\cite{mimno07topicbased,charlin13tpms,wieting2019simple,cohan2020specter,mysore2023editable,singh2022scirepeval}; see~\cite[Section 3]{shah2022challenges} for a more extensive survey. In this work we focus on the SPECTER model~\cite{cohan2020specter}, described in Section~\ref{sec:text_sim}, due to its widespread use. We note that a second version of SPECTER is also available~\cite{singh2022scirepeval}. In preliminary experiments, we found that SPECTER version 2 has similar vulnerabilities as the original SPECTER model, and we focus on SPECTER since this is still the most widely used model in ML/AI conferences. 

\paragraph{\bf Collusion rings} The problem of collusion rings was first uncovered in computer science in the field of computer architecture~\cite{vijaykumar2020asplos} and subsequently also in ML/AI conferences~\cite{littman2021collusion}. Among the various components of the similarity score computation, research on addressing the problem of collusion rings has primarily focused on bidding~\cite{wu2021making,jecmen2022tradeoffs,jecmen2022dataset, jecmen2024detection,leyton2024matching}. In practice, investigations into collusions~\cite{ICLR2024chairs} have also concentrated on bidding, and most efforts to mitigate collusion rings~\cite{leyton2024matching} have similarly focused on bidding. Keywords and subject areas are also considered susceptible~\cite{ailamaki2019sigmod}. Some other work impart additional constraints~\cite{guo2018k,boehmer2022combating} or randomness in the reviewer assignment~\cite{jecmen2020mitigating,xu2024one}.

\paragraph{\bf Text matching and collusion rings}
We review several studies that address text matching in the context of collusion rings. Previous research in the security community~\cite{markwood2018mirage, tran2019pdfphantom} has demonstrated successful malicious attacks that manipulate fonts embedded in the PDF (Portable Document Format) of a submitted paper to ensure that a colluding reviewer is assigned to evaluate it. These attacks exploit the fact that automated text similarity tools depend on PDF parsers, allowing malicious authors to conceal text in their submissions that remains invisible to human readers but is detected by software. However, these tactics lack plausible deniability, meaning anyone caught tampering with their PDF submission risks damaging their career and reputation.

\paragraph{\bf Text-level attacks and collusion rings} 
Previous work demonstrated the possibility of text-level attacks through editing the bibliography, replacing synonyms, introducing spelling mistakes, and using language models to incorporate keywords \cite{eisenhofer2023no}. They showed that malicious authors can successfully select and remove reviewers by adding and removing topical keywords in their paper submissions. \cite{eisenhofer2023no} is by far the most related to this paper, and we discuss the differences between their work and ours in much more detail in the remainder of this section.
\begin{enumerate}
    \item \textbf{Setting} While the demonstration of manipulation in their work is on a small scale of 165 reviewers and 32 papers, our work considers a scale two orders of magnitude larger with 7,900 reviewers and 3,218 papers. We also consider attacks when the original ranking of the reviewer for the target paper is in the range of 20 to 1001, as compared to the original ranking of 6 to 10 considered in their work. An additional assumption made in their work is that the attackers know the program committee beforehand. This assumption is not suitable for our setting (ML/AI venues), so we also dispose of this assumption. Finally, their work focuses only on modifying the paper submission, but our work additionally considers adversarial archive curation by the reviewer as another attack surface. 
    \item \textbf{Text matching model} Their work considers a Latent Dirichlet Allocation (LDA) based topic modeling algorithm used in the Toronto Paper Matching System (TPMS) for text matching. In our work, we instead consider a modern neural network embedding based text similarity model~\cite{cohan2020specter}. The model we consider is much less interpretable than LDA-based models, and it is more widely used at ML/AI venues. The paper~\cite{eisenhofer2023no} further investigates the possibility of black box attacks without access to the reviewer assignment model. However, in our scenario, we assume that anyone has access to the open-sourced SPECTER model. 

    \item \textbf{Evaluation} The manipulations in their work involve PDF tricks, without which they achieve relatively low success rates ($<70\%$ even with aggressive modifications to the paper); in contrast, we show that assignment manipulation purely through text manipulations is possible with high success rates ($\sim 92\%$). In our work, we focus on investigating the possibility of a paper submission getting assigned to a colluding reviewer, without considering adding or removing multiple reviewers at once. In their ablation, they demonstrated that it is possible to select all five reviewers adversarially using PDF manipulations to add or remove a median of 5,968 words from the submissions. Finally, while both papers include human subject experiments to test for the detectability of adversarial papers, we collected 116 samples, which is considerably more than the 21 samples they collected.
    
    \item \textbf{Publicly released adversarial samples} While \cite{cohan2020specter} has only released three adversarial samples, we publicly release all our adversarial samples (1000+ total samples) for complete transparency. 
\end{enumerate}

\section{Attack Procedure} \label{sec:procedure}

\begin{figure}[tb]
\centering
\begin{subfigure}[tb]{0.5\linewidth}
    \centering\includegraphics[trim={5cm 4.7cm 34cm 11.7cm},clip,width=0.85\linewidth]{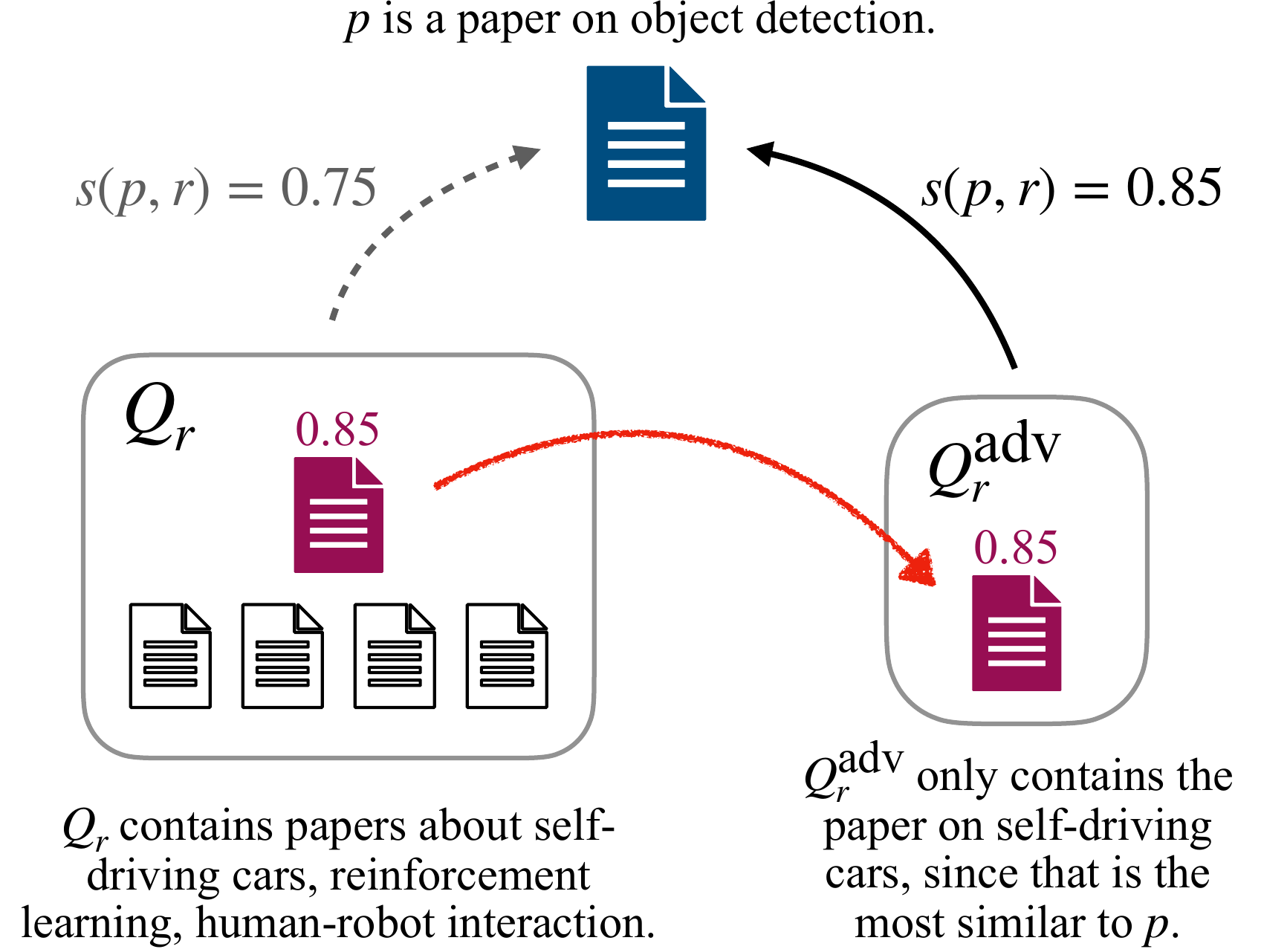}
    \caption{The colluding reviewer adversarially curates their archive to only contain one of their past papers that is the most similar to the paper $p$ they want to get assigned.} \label{subfig:reviewer_action}
\end{subfigure}
\hfill
\begin{subfigure}[tb]{0.45\linewidth}
    \centering\includegraphics[trim={37cm 4.2cm 5cm 11.2cm},clip,width=0.75\linewidth]{figures_pdf/reviewer-author.pdf}
    \caption{The colluding author adversarially modifies their abstract to be more similar to the paper in the colluding reviewer's curated archive.} \label{subfig:author_action}
\end{subfigure}
\caption{An illustrated example of the attack procedure. \ref{subfig:reviewer_action} shows the reviewer's action; \ref{subfig:author_action} shows the author's action.}   
\label{fig:attack}
\end{figure}

For any colluding paper-reviewer pair $\PaperReviewerPair$, the attack's goal is to have $\Reviewer$ be assigned to $\Paper$ for peer review at a conference. We begin by detailing the reviewer's adversarial action in Section \ref{sec:reviewer}; then, we detail the author's adversarial action in Section \ref{sec:author}. Figure \ref{fig:attack} is an overview of the procedure.

\subsection{Adversarial Archive Curation by the Colluding Reviewer} \label{sec:reviewer}

Peer review assignment systems often allow reviewers to curate their own archives for text-matching. 
Some conferences even encourage this practice, starting with an empty archive that reviewers can then populate with any papers they choose. 
Platforms like OpenReview and the Toronto Paper Matching System (TPMS) \cite{charlin2013toronto} allow reviewers to manually add arbitrary publications by specifying titles and abstracts, or remove articles from their current profiles. While the ability to manually add arbitrary publications can amplify the effectiveness of attacks, our analysis focuses on the more common case where reviewers select papers from their existing profiles.

A colluding reviewer can exploit this process by constructing an adversarial archive, $\ArchiveAdv{\Reviewer}$, that includes only papers from their archive $\Archive{\Reviewer}$ which are highly similar to the target paper $\Paper$. With max pooling, a commonly used aggregation strategy (Section~\ref{sec:text_sim}), the reviewer only needs to ensure that their archive retains the paper most similar to $\Paper$. Since max pooling considers only the highest similarity score, the presence of one highly similar paper is sufficient to bias the assignment, making this method particularly vulnerable.

Average pooling could be more robust to such outlier high similarity scores, but the colluding reviewer can still break average pooling by curating their archive to keep only one paper that has the highest similarity to $\Paper$, 
\begin{equation} \label{eq:adversarial_archive}
    \ArchiveAdv{\Reviewer} = \left \{ \text{a random paper in }  \operatorname*{argmax}_{\PastPaper \in \Archive{\Reviewer}} \frac{\Embedding{\Paper} \cdot \Embedding{\PastPaper}}{\norm{\Embedding{\Paper}} \norm{\Embedding{\PastPaper}}} \right \} ,
\end{equation}
with ties broken uniformly at random, where $\Embedding{\Paper} \in \mathbb{R}^k$ is the SPECTER embeddings defined in Section \ref{sec:text_sim}.

For simplicity, the majority of our analyses focuses on the setting where the colluding reviewer has a single paper in their archive. Under max pooling (with arbitrary length archives), the similarity score obtained by the colluding reviewer for the target paper is greater than or equal to the similarity score obtained under our single paper setting. Thus, while these analyses of a single paper are not identical to max pooling, they give indications of attack potential under both max pooling and under reviewer-driven sub-selection under average pooling. We rigorously compare the two settings in 
Section~\ref{sec:pooling} and find that max pooling is indeed more vulnerable. 

\subsection{Adversarial Abstract Modifications by the Colluding Author} \label{sec:author}

\paragraph{Notation.} 
Recall that only titles and abstracts are used to compute similarity scores between papers, and reviewers can curate their archives adversarially. 
In this section, we will expand notation for paper $\Paper$ to have corresponding $\Title$ for the title and $\Abstract$ for the abstract as we will manipulate those, and we will explicitly denote the reviewer archive. When the archive has only one paper, max and average pooling are the same and we drop the subscript for brevity in notation. Hence, we use $\SimilarityAbstract{\Title}{\Abstract}{\Archive{\Reviewer}}$ to denote paper-reviewer similarity below.

\paragraph{Key idea.} To have a paper $\Paper$ assigned to the colluding reviewer $\Reviewer$ for peer review, the key idea behind the authors' steps is to construct an adversarial abstract $\PaperAbstractAdv{\Paper}$ to increase the similarity $\SimilarityAbstract{\Title}{\PaperAbstractAdv{\Paper}}{\ArchiveAdv{\Reviewer}}$ from $\SimilarityAbstract{\Title}{\Abstract}{\ArchiveAdv{\Reviewer}}$ while maintaining consistency and coherence to avoid arousing suspicion. 

We discover two simple abstract modification operations that effectively increase the similarity. The first is \IncludeRelatedWork{} which involves adding background or filler sentences related to the main ideas of papers in $\ArchiveAdv{\Reviewer}$. The second is \InsertKeywords{} which inserts ``keywords'' that target SPECTER similarity to increase $\SimilarityAbstract{\Title}{\AbstractAdv}{\ArchiveAdv{\Reviewer}}$ even if the keywords do not necessarily seem important to humans. This operation is inspired by works in adversarial robustness that show how small unexpected and unintelligible changes can break machine learning classifiers~\cite{ebrahimi2017hotflip,jin2020bert,li2020bert}.

For each operation, we provide a description of the operation and an example modified abstract. Both operations employ Large Language Models (LLMs) and have two modes --- \textit{human-in-the-loop} and \textit{fully automatic}. The \textit{human-in-the-loop} mode is the way we expect colluding authors would modify their abstracts in reality, as human oversight can ensure there are no suspicious artifacts in the adversarial abstract. For several experiments in this paper, we use the \textit{fully automatic} mode for scalability, but we also simulate and evaluate \textit{human-in-the-loop} attacks at a feasible scale. The LLM model we use in this paper is the OpenAI gpt-4-0125-preview model.

Since adversarial abstract modification is a multi-step process, we also propose an \textbf{optional} early stopping check to stop further modifications if the attack is already envisaged to be successful. In one experiment, we use a realistic early stopping heuristic derived from past conference's data: \textit{if the colluding reviewer is the most-similar reviewer for the colluding paper amongst all NeurIPS 2022 reviewers, no further abstract modifications are made}. See Appendix \ref{appx:algorithms} for details.

\subsection*{The \IncludeRelatedWork{} operation } 
The goal of \IncludeRelatedWork{} is to modify the abstract $\Abstract$ in order to increase the SPECTER similarity between paper $\Paper$ and reviewer $\Reviewer$, by adding background or filler sentences related to the central themes of papers in $\ArchiveAdv{\Reviewer}$. We note that the resulting modified abstract $\PaperAbstractAdv{\Paper}$ may become inconsistent with its paper $\Paper$ if the main ideas in the abstract have been changed entirely. Our key observation is that SPECTER similarities increase when abstracts share overlapping themes, even when those themes are central to papers in $\ArchiveAdv{}$ but only referenced as background information in $\AbstractAdv$. 
This means that \IncludeRelatedWork{} can produce $\AbstractAdv{}$ by including themes from $\ArchiveAdv{}$ to increase $\SimilarityAbstract{\Title}{\AbstractAdv}{\ArchiveAdv{}}$ without violating the coherence and consistency constraints.
An example of $\IncludeRelatedWork{}$ can be found in Figure $\ref{fig:include_themes_example}$. 

\begin{figure}[tb]
    \centering
    \fbox{
    \begin{minipage}{0.95\linewidth}
     \textbf{\IncludeRelatedWork{} Example:} {\sethlcolor{Goldenrod}\hl{Multimodal language models have shown promise in AI applications like robotics, where these models enable scalable approaches for learning open-world object-goal navigation -- the task of asking a virtual robot agent to find any instance of an object in an unexplored environment (e.g., ``find a sink'').}} In this work, we propose a new method to fuse the embedding space of frozen text-only large language models (LLMs) and pre-trained image encoder and decoder models, by mapping between their embedding spaces. Our model demonstrates a wide suite of multimodal capabilities: image retrieval, novel image generation, and multimodal dialogue\ldots
    \end{minipage}
    }
    \caption{An example of \IncludeRelatedWork{} modifying the abstract. Goal navigation is an important theme in the adversarial archive $\ArchiveAdv{}$ of the reviewer $\Reviewer$. The modification (highlighted sentence) adds robot goal navigation as a motivating example for further improvements to multimodal models. The resulting abstract $\AbstractAdv$ remains coherent and consistent with the paper's focus on text and image embedding alignment, but has increased similarity to the target adversarial archive $\SimilarityAbstract{\Title}{\AbstractAdv}{\ArchiveAdv{}}$. The \IncludeRelatedWork{} changes shown in this example, alongside adversarial reviewer archive curation, increase the reviewer's similarity to the paper from being 101st most-similar to become the \emph{most} similar amongst all reviewers at the NeurIPS 2023 conference.}
    \label{fig:include_themes_example}
\end{figure}

The implementation of the \HumanMode{} and \AutoMode{} modes differ in whether human supervision is used to ensure the coherence and consistency of the modified abstract $\AbstractAdv$. In the \textit{human-in-the-loop} mode, modified abstracts $\AbstractAdv$ are created by a human (potentially with the help of a LLM), and the human can make incremental edits to $\AbstractAdv$ to ensure coherence and consistency. On the other hand, the \textit{fully automatic} mode only uses a LLM to generate the $\AbstractAdv$ with a prompt that emphasizes coherence and consistency, and it does not allow further edits to $\AbstractAdv$ once they are generated. 

In addition, the implementation for both \HumanMode{} and \AutoMode{} modes involves generating $\Regenerations$ different versions of modified abstracts. Since generating and modifying abstracts is stochastic (due to the stochasticity of LLM outputs and manual edits), we keep only the attempt that is the most similar to the colluding reviewer $\Reviewer$. 
We tune and report the choice of $\Regenerations$ for our experiments in Section \ref{sec:exp_setup}, and the formal algorithm is Algorithm \ref{alg:include_theme} in Appendix \ref{appx:include_related_work_alg}.

\subsection*{The \InsertKeywords{} operation} 

\begin{figure}[tb]
    \centering
    \fbox{
    \begin{minipage}{0.95\linewidth}
     \textbf{Example:} 
     With more real-world machine \sethlcolor{Goldenrod}\hl{learning} applications, the importance of safeguarding data privacy of \sethlcolor{Goldenrod}\hl{labels} and \sethlcolor{Goldenrod}\hl{features} has increased \sethlcolor{Goldenrod}\hl{\textbf{manifold}}. 
     Per-example gradient clipping is a key algorithmic step that enables practical differential private (DP) training for deep learning models. The choice of clipping threshold $R$, however, is vital for avoiding high training loss \sethlcolor{Goldenrod}\hl{\textbf{discrepancy}} and achieving high accuracy under DP. 
     Not only does it serve as a clipping threshold, but the \sethlcolor{Goldenrod}{Gaussian} noises added are also dependent on $R$. 
     We propose an easy-to-use replacement, called automatic clipping, that eliminates the need to tune $R$ for any DP optimizers, including DP-SGD, DP-Adam, DP-LAMB and many others. The automatic variants are as private and computationally efficient as existing DP optimizers, but require no DP-specific hyperparameters and thus improve DP training \sethlcolor{Goldenrod}\hl{\textbf{manifold}}. Our proposed method is as amenable as the standard non-private training \sethlcolor{Goldenrod}\hl{\textbf{flow}}\ldots
    \end{minipage}
    }
    \caption{An example of \InsertKeywords{} modifications. The paper in the adversarial archive $\ArchiveAdv{}$ proposes mitigating label scarcity in transfer learning with data augmentation. 
    Some inserted keywords (highlighted) match their original meanings from the paper in $\ArchiveAdv{}$, while others (highlighted + bolded) are adapted to common English.
    The \InsertKeywords{} changes to this abstract, alongside adversarial reviewer archive curation, increase the reviewer's similarity to the paper from being 101st most-similar to 3rd most-similar amongst all reviewers at the NeurIPS 2023 conference.}
    \label{fig:insert_keywords_example}
\end{figure}

Next, \InsertKeywords{} adds specific \textit{keywords} to the abstract. These keywords may not seem obviously important to humans, but they can increase $\SimilarityAbstract{\Title}{\AbstractAdv}{\ArchiveAdv{}}$ if added into the adversarial abstract. This phenomenon aligns with findings in the adversarial examples literature; the outputs of neural networks can be brittle to the insertion of certain keywords or tokens into the input. Furthermore, we find that the similarities can increase even when the keywords are used \textit{under different meanings and with different parts of speech} across abstracts. This allows \InsertKeywords{} to insert technical keywords from $\ArchiveAdv{}$ into the $\AbstractAdv$ without introducing unrelated or suspicious concepts. 
For example, ``transfer learning'' might be simplified to ``transfer,'' a common English term. To ensure coherence and grammatical correctness, \InsertKeywords{} may also adjust text around the inserted keywords (e.g. insert a phrase that contains the keyword).

We present an example of $\InsertKeywords{}$ in Figure \ref{fig:insert_keywords_example}. In this example, the paper in adversarial archive $\ArchiveAdv{}$ proposes mitigating label scarcity in transfer learning with data augmentation, in particular using gradient flow methods to the minimize maximum mean discrepancy loss on the feature-Gaussian manifold. 
We also propose the \FindKeywords{} subroutine, which is described later, to suggests keywords from $\ArchiveAdv{}$ to add to the abstract $\Abstract$, including repeated words if they can further increase similarity.

Some keywords from the reviewer's archive $\ArchiveAdv{}$ carry meanings directly related to the abstract $\Abstract$. In Figure \ref{fig:insert_keywords_example}, the keyword `learning' is added as `machine learning';
`feature-label' is broken up and added as `labels and features'; `feature-Gaussian' is inserted only as `Gaussian'. For these keywords, the inserted variations match their meanings in the reviewer's archive.
However, other keywords with technical meanings unrelated to this paper $\Paper$, such as `manifold', `discrepancy', and `flow', are adapted to common English (`manifold' as ``a great deal'', `discrepancy' as ``difference'', and `flow' referring to training pipelines). Still, there can be unrelated technical keywords like `Riemannian' and `optimum' that do not have suitable common English meanings for abstract $\Abstract$, so they would not be inserted.

Now, we describe the implementation details of the \InsertKeywords{} operation.
The process iteratively searches for $\Alternations$ batches of $\Keywords$ keywords, inserting each batch before searching for the next. We propose the \FindKeywords{} subroutine (Algorithm \ref{alg:keywords} in Appendix \ref{appx:insert_keywords_alg}) to greedily choose keywords that increase the similarity $\SimilarityAbstract{\Title}{\AbstractAdv}{\ArchiveAdv{}}$.
Since $\FindKeywords{}$ is a greedy algorithm, the alternation between finding and inserting each batch of keywords can take into account the new $\AbstractAdv$ when finding the next batch of keywords. 
The two hyperparameters, $\Alternations$ and $\Keywords$, in \InsertKeywords{} are tuned and reported in Section \ref{sec:exp_setup}. 
The formal algorithm of \InsertKeywords{} operation is detailed in Algorithm \ref{alg:insert_keywords} in Appendix \ref{appx:insert_keywords_alg}.

\InsertKeywords{} also has \textit{human-in-the-loop} and \textit{fully automatic} modes (detailed in Algorithm \ref{alg:insert_keywords}). The differences between the two modes lie in how the keywords are inserted. In the \textit{human-in-the-loop} mode, the human adds the keywords into $\AbstractAdv$ one by one. In addition, the human creates up to five drafts of different ways to add each keyword and keeps the draft with the maximum similarity. 
Since manually enforcing coherence and consistency constraints greatly limits where each keyword can be inserted in $\AbstractAdv$, we find that taking the maximum of multiple drafts is helpful, especially when the similarity is sensitive to the position each keyword is inserted at.
In the \textit{fully automatic} mode, LLM inserts a whole batch of keywords each time, and the LLM is prompted to follow coherence and consistency constraints, but there are no human supervision to ensure them.

\section{Results}
In this section, we present the results of our experiments. 
We explain the experimental setup in Section \ref{sec:exp_setup}.
In Section \ref{sec:manual_result}, we simulate realistic human-in-the-loop attack scenarios and evaluate the attack effectiveness on 25 randomly selected (paper, reviewer) pairs of colluders.
Then, we conduct larger-scale experiments using automatic abstract modifications and investigate the attack effectiveness even when colluding reviewers have low natural rankings (Section \ref{sec:automatic_result}). 

We also discuss scenarios where the attack success rates can be reduced in Sections \ref{sec:arclen} and \ref{sec:pooling}. Furthermore, we discuss the potential usage of publicly released reviewer pool data for attack refinement in Section \ref{sec:correlation}. Finally, in Section \ref{sec:human_subject_experiment} we discuss a human subject experiment testing for the suspiciousness and detectability of the modified abstracts. All adversarial abstracts generated for all experiments are publicly downloadable from Zenodo at \url{https://doi.org/10.5281/zenodo.15588237}. 

\subsection{\textbf{Experiment Setup}} \label{sec:exp_setup}
To evaluate the attack procedure, we download a dataset of reviewer archives and papers from the Neural Information Processing Systems (NeurIPS) 2023 conference. We also consider the previous edition, NeurIPS 2022, as a publicly available ``prior'' conference to develop the attack algorithm. 
Our setup simulates the real-world scenario when colluders only have access to the data of prior conferences before submitting to a new conference.

\paragraph{\textbf{Dataset}} We download all accepted papers at the NeurIPS 2023 venue using the OpenReview API (\url{https://api2.openreview.net}). 
We download the names of all reviewers at NeurIPS 2023 (\url{https://neurips.cc/Conferences/2023/ProgramCommittee}) and search for OpenReview profiles that match the names of each reviewer. 
When curating the reviewer pool, we discard some reviewers if (1) there are multiple profiles that match the name of a reviewer or if (2) the reviewer has no public publications on their OpenReview profile. In this manner, we obtain 3,218 papers and 7,900 reviewers for the experiments. Following the same procedure, we also curate a NeurIPS 2022 dataset with 2,671 papers and 6,634 reviewers.
We only download and use the paper metadata where Creative Commons Public Domain Dedication (CC0 1.0) apply (\url{https://openreview.net/legal/terms}).

\paragraph{\textbf{Similarity rankings}} Our evaluations are based on the competition rankings of reviewer similarities, where reviewers that have equal similarities receive the same ranking number, and then a gap is left in the ranking numbers (e.g. ``1,2,2,4''). For any paper $\Paper$ and reviewer $\Reviewer$ at a conference, we define the \textit{natural ranking} of ($\Paper$, $\Reviewer$) as the competition ranking of $\Reviewer$ when all reviewers at the conference are ranked by their similarity scores with the paper $\Paper$, without manipulations. We use the term \textit{manipulated ranking} for the colluding reviewer's competition ranking by similarity to the colluding paper as a result of their manipulations. Appendix \ref{appx:natural_sim_and_rankings} relates natural rankings to absolute similarity. 

\paragraph{\textbf{Evaluation samples}}
From the curated NeurIPS 2023 papers and reviewers, for each paper $\Paper$, we find reviewers (there may be multiple since competition ranking is used) with natural rankings of 101. If no reviewer ranked 101$^{st}$ for a paper due to ties, we find the reviewers who have the next rank after 100. From all (paper $\Paper$, reviewer $\Reviewer$) pairs collected this way, we randomly sample without replacement a number of $\PaperReviewerPair{}$ pairs to act as colluders running the proposed attack algorithm. We evaluate the attack efficacy on these colluding paper-reviewer pairs.
OpenReview sets the similarity score to be $0$ if reviewer $r$ has a natural ranking greater than 100 for paper $\Paper$, so it would be highly unlikely for $\Reviewer$ to be assigned to review $\Paper$ naturally. Similarily, we also sample $\PaperReviewerPair{}$ pairs with natural rankings of 20, 501 and 1001 for evaluation. We specify exact sample sizes in each section.

\paragraph{\textbf{Evaluation metrics}}
We evaluate the effectiveness of our attack by measuring the top-1, top-3, and top-5 success rates in NeurIPS 2023, where a top-N success rate is defined as the fraction of times the attack successfully increases the colluding reviewer's manipulated ranking to be top-N for the colluding paper. We study these success rates because most papers are assigned to 3--6 reviewers at conferences \cite{shah2022challenges}.

\paragraph{\textbf{Attack algorithm}} The colluding reviewer constructs $\ArchiveAdv{}$ by selecting the most similar paper to $\Paper$ from the default archive $\Archive{}$ of up to 10 most-recent papers they have authored (Section \ref{sec:reviewer}). In addition, the colluding author modifies their adversarial abstract $\AbstractAdv$ to be more similar to $\ArchiveAdv{}$ (Section \ref{sec:author}). We investigate \textit{human-in-the-loop} attack efficacy with early stopping in Section \ref{sec:manual_result}. In subsequent sections, we investigate \AutoMode{} attack results and do not use early stopping. The \AutoMode{} mode does not require human supervision, which makes large-scale evaluations and further ablations feasible.

\paragraph{\textbf{Attack budgets (hyperparameters)}}
There are three hyperparameters $\Regenerations,\Alternations,\Keywords$ we use to define the attack budget. In \IncludeRelatedWork{}, $\Regenerations$ stands for the number of $\AbstractAdv$ versions created before selecting the most similar version. In \InsertKeywords{}, $\Alternations$ is the number of batches of keywords to insert, and $\Keywords$ is the maximum number of keywords in each batch. We explore the attack success rates under different  $\Regenerations,\Alternations,\Keywords$ with the NeurIPS 2022 dataset and select the highest performing combination $\Regenerations=5$, $\Alternations=2$, $\Keywords=5$. In Appendix \ref{sec:budgets}, we present an investigation into different choices of hyperparameters $\Regenerations, \Alternations, \Keywords$ on the NeurIPS 2023 dataset.

\paragraph{\textbf{LLM used}} In our experiments, we use the OpenAI gpt-4-0125-preview model with temperature 1 for abstract modifications in both \HumanMode{} and \AutoMode{} modes of \IncludeRelatedWork{} and \InsertKeywords{} operations.

\subsection{\textit{Human-in-the-loop} Mode Attack Results} \label{sec:manual_result}
We perform 25 attacks by constructing the $\ArchiveAdv{}$ and modifying $\AbstractAdv$ in the more realistic \textit{human-in-the-loop} mode. We randomly sample \PaperReviewerPair{} pairs with natural rankings of 101, and we keep the first 25 samples with paper topics we are familiar enough with to judge the coherence and consistency of the modified abstracts. Firstly, we use the method described in Section \ref{sec:reviewer} to construct $\ArchiveAdv{}$. For the abstract modifications, we simulate what colluding authors would do when abstracts are modified with human involvement. The \textit{human-in-the-loop} implementations of \IncludeRelatedWork{} and \InsertKeywords{} are described in Appendix \ref{appx:algorithms} Algorithm \ref{alg:include_theme} and Algorithm \ref{alg:insert_keywords}, respectively. In addition, we do early stopping checks to prevent abstracts from being modified more than necessary. The exact early stopping heuristics used can be found in Section \ref{sec:author} and Appendix \ref{appx:algorithms}.

For the \textit{human-in-the-loop} attack, we find it helpful to increase the default attack budget $\Regenerations$, which is the number $\AbstractAdv$ versions generated in \IncludeRelatedWork{}, from 5 to 10. This is because the consistency and coherence constraints are enforced much more strictly in the \HumanMode{} mode. 
In cases when LLM outputs a version that is similar to previous versions or does not increase the similarity to $\Reviewer$ meaningfully, we skip that version and move on to the next version to save editing time. 

In addition to increasing $\Regenerations$, we allow up to 3 sentences about the colluding reviewer's archive $\Archive{\Reviewer}$ to be added to the abstract during \IncludeRelatedWork{}, instead of the 1 sentence constraint in the fully automatic mode. This is a realistic change because coherence and consistency are enforced manually here, and additional sentences often improve abstract flow by allowing better transition sentences. As for \FindKeywords{} and \InsertKeywords{}, we choose the default values of $\Keywords=5$ and $\Alternations=2$. In fact, because of the early stopping, most abstracts have less than 10 keywords added.

\begin{table}[tb]
\caption{Attack success rates in \textit{human-in-the-loop} mode with early stopping.}
\centering
\renewcommand{\arraystretch}{1.2}
\begin{tabulary}{\linewidth}{@{}CCCCC@{}}
 \toprule
 & \multicolumn{3}{c}{Attack Success Rates ($\pm$ SE)} & Manipulated Ranking\\ \cmidrule(lr){2-4} \cmidrule(lr){5-5}
 \mbox{Natural} & \multirow{2}*{Top-1} & \multirow{2}*{Top-3} & \multirow{2}*{Top-5} & Mean\\
 \mbox{Rankings}& & & & 95\% CI \\ \midrule
 
 \multirow{2}*{101} & \multirow{2}*{\SI{76 \pm 9}{\percent}} & \multirow{2}*{\SI{92 \pm 6}{\percent}} & \multirow{2}*{\SI{92 \pm 6}{\percent}} & 2.08 \\ 
 & & & & [0.97, 3.19] \\ 
 \bottomrule
\end{tabulary}
\label{tab:manual_success}
\end{table}

We report the attack success rates in Table \ref{tab:manual_success}. 
We find that the human-in-the-loop attacks with early stopping can successfully increase the colluding reviewers' manipulated rankings in most cases. 
Even when coherence and consistency constraints are manually enforced, the proposed attack procedure still have high success rates.

\subsection{\textit{Fully Automatic} Mode Attack Results} \label{sec:automatic_result}

In the remainder of this paper, we investigate the success rates of our attack under \AutoMode{} mode abstract manipulation without early stopping. The \textit{fully automatic} implementation of \IncludeRelatedWork{} and \InsertKeywords{} can be found in Algorithm \ref{alg:include_theme} and Algorithm \ref{alg:insert_keywords}, respectively. Unlike the \HumanMode{} results in Section~\ref{sec:manual_result}, the \AutoMode{} mode involves no human supervision of the LLM outputs. Our motivation for using the \AutoMode{} mode is to enable large-scale, reproducible evaluations.

We randomly sample without replacement 300 (paper $\Paper$, reviewer $\Reviewer$) pairs where $\Reviewer$ has a natural ranking of 101 for $\Paper$. For the natural ranking of 20, 501 and 1001, we sample 100 $\PaperReviewerPair$ pairs. (The larger sample size for the 101st scenario is due to a request from a reviewer of this paper.)
Afterwards, we perform the attack procedure for each $\PaperReviewerPair$ pair by curating the adversarial archive $\ArchiveAdv{}$ (Section~\ref{sec:reviewer}) and then performing \AutoMode{} mode abstract modifications (Section~\ref{sec:author}).

We enumerate the attack success rates in Table \ref{tab:automatic_success}. We find that the success rates are generally high. When the natural ranking is 20, the attack success rates are the highest since the colluding $\PaperReviewerPair$ pair is naturally highly similar. When the natural ranking is 101, the proposed attack procedure leads to a top-5 attack success rate of 93\%. Even when the natural ranking is 1001, the top-1 attack success rate is 48\%. These results highlight that colluding $\PaperReviewerPair$ pairs can successfully manipulate reviewer assignments, whether they work on similar topics or not. 
Appendix~\ref{appx:success_stratified} further explore the relationships between success rates and absolute similarity.

In the following subsections, we present further insights into the efficacy of various attacks components.

\begin{table}[h]
\caption{Attack success rates in \textit{fully automatic} mode for colluding reviewers with natural rankings of 20, 101, 501, and 1001.}
\centering
\renewcommand{\arraystretch}{1.2}
\begin{tabulary}{\linewidth}{@{}CCCCC@{}}
 \toprule
 & \multicolumn{3}{c}{Attack Success Rates ($\pm$ SE)} & Manipulated Rankings\\ \cmidrule(lr){2-4} \cmidrule(lr){5-5}
 \mbox{Natural}& \multirow{2}*{Top-1} & \multirow{2}*{Top-3} & \multirow{2}*{Top-5} & Mean\\
 \mbox{Rankings}& & & & 95\% CI \\ \midrule

  \multirow{2}*{20} & \multirow{2}*{\SI{90 \pm 3}{\percent}} & \multirow{2}*{\SI{96 \pm 2}{\percent}} & \multirow{2}*{\SI{98 \pm 1}{\percent}} & 1.28 \\ 
 & & & & [1.06, 1.50] \\ \cmidrule(r){1-5}

 \multirow{2}*{101} & \multirow{2}*{\SI{74 \pm 3}{\percent}} & \multirow{2}*{\SI{89 \pm 2}{\percent}} & \multirow{2}*{\SI{93 \pm 2}{\percent}} & 2.22 \\ 
 & & & & [1.80, 2.64] \\ \cmidrule(r){1-5}
 
 \multirow{2}*{501} & \multirow{2}*{\SI{60 \pm 5}{\percent}} & \multirow{2}*{\SI{76 \pm 4}{\percent}} & \multirow{2}*{\SI{83 \pm 4}{\percent}} & 6.58 \\ 
 & & & & [3.47, 9.69] \\ \cmidrule(r){1-5}

 \multirow{2}*{1001} & \multirow{2}*{\SI{48 \pm 5}{\percent}} & \multirow{2}*{\SI{63 \pm 5}{\percent}} & \multirow{2}*{\SI{67 \pm 5}{\percent}} & 15.68 \\ 
 & & & & [6.82, 24.54] \\ 
 \bottomrule
 
\end{tabulary}
\label{tab:automatic_success}
\end{table}

\subsection{Lower Limits on Reviewer Archive Length} \label{sec:arclen}

Currently, in most venues, reviewers are allowed -- and frequently even encouraged -- to curate their profiles with the goal of allowing reviewers to keep only relevant and representative papers. However, an adversary can misuse the current system to curate their archive, keeping their colluding author's submission in mind. In this section, we investigate the specific vulnerability of this policy under average (mean) pooling and investigate potential defenses.

What is the effect of simply curating the adversarial archive $\ArchiveAdv{}$ as described in Section \ref{sec:reviewer}, without any manipulation of abstracts by authors ($\AbstractAdv{}=\Abstract$)? For colluding $\PaperReviewerPair{}$ pairs with natural rankings of 101, we find that 30\% of the samples have manipulated rankings being within the top-3 from adversarial reviewer archive curation alone---see Table \ref{tab:no_author}. In addition, the mean manipulated ranking is 24.59, which is much improved compared to the natural ranking of 101. This is in fact a significant vulnerability, and reviewer archive curation \emph{alone} is a serious threat to automated reviewer assignments.

\begin{figure}[h]
    \centering
    \includegraphics[scale=0.4]{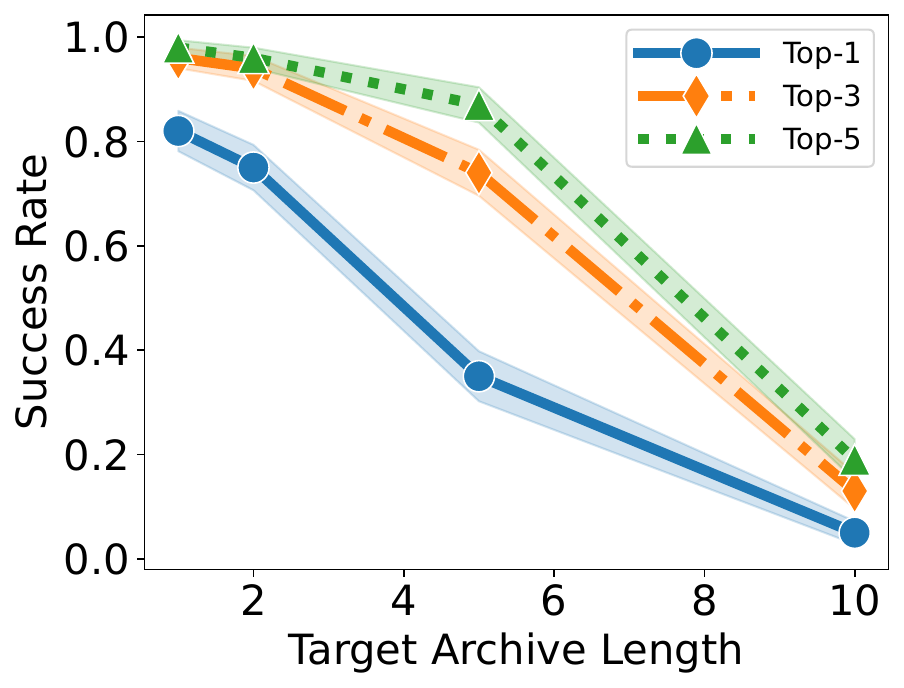}
    \caption{Attack success rates when the colluding reviewers have to keep 1, 2, 5, 10 papers in the adversarial archive $\lvert \ArchiveAdv{} \rvert$. Success rates drop when colluding reviewers must keep more papers in their archive. The shaded bands represent standard errors of the mean.}
    \label{fig:arclen}
\end{figure}

A possible defense to our attack could be imposing a lower limit on the number of publications each reviewer has to keep in their archive. To investigate such a defense, we randomly sample without replacement 100 $\PaperReviewerPair$ pairs to act as colluders with natural rankings of 101 and where the reviewer $\Reviewer$ has at least ten publications. For the curation of $\ArchiveAdv{}$, we consider four different scenarios where the reviewers keep the 1, 2, 5, or 10 most-similar publications in $\ArchiveAdv{}$. Afterwards, we run automatic abstract modification and evaluate the success rates. Figure \ref{fig:arclen} shows that attack success rates decreases with $\lvert \ArchiveAdv{} \rvert$, meaning that imposing a high lower limit on the reviewer' archive lengths can effectively decrease the proposed attack's success rates. However, there is a trade-off here, since honest reviewers may actually want to update their profiles to reflect their most current research interests.

\begin{table}[h]
\caption{Attack performance with reviewer action but without abstract manipulation.}
\centering
\renewcommand{\arraystretch}{1.2}
\begin{tabulary}{\linewidth}{@{}CCCCC@{}}
 \toprule
 & \multicolumn{3}{c}{Attack Success Rates ($\pm$ SE)} & Manipulated Rankings\\ \cmidrule(lr){2-4} \cmidrule(lr){5-5}
 \mbox{Natural} & \multirow{2}*{Top-1} & \multirow{2}*{Top-3} & \multirow{2}*{Top-5} & Mean\\
 \mbox{Rankings} & & & & 95\% CI \\ \midrule

 \multirow{2}*{101} & \multirow{2}*{\SI{15 \pm 2}{\percent}} & \multirow{2}*{\SI{30 \pm 3}{\percent}} & \multirow{2}*{\SI{41 \pm 3}{\percent}} & 22.80 \\ 
 & & & & [19.16, 26.45] \\ 
 \bottomrule
 
\end{tabulary}
\label{tab:no_author}
\end{table}

\subsection{Maximum versus Average Similarity} \label{sec:pooling}

\begin{table}[tb]
\caption{Attack performances using average or max pooling in paper-reviewer similarity calculation (natural ranking 101).}
\centering
\renewcommand{\arraystretch}{1.2}
\begin{tabulary}{\linewidth}{@{}CCCCC@{}}
 \toprule
 & \multicolumn{3}{c}{Attack Success Rates ($\pm$ SE)} & Manipulated Rankings\\ \cmidrule(lr){2-4} \cmidrule(lr){5-5}
 \mbox{Aggregati-} & \multirow{2}*{Top-1} & \multirow{2}*{Top-3} & \multirow{2}*{Top-5} & Mean\\
 \mbox{on Method} & & & & 95\% CI \\ \midrule

 \multirow{2}*{Average} & \multirow{2}*{\SI{13 \pm 3}{\percent}} & \multirow{2}*{\SI{24 \pm 4}{\percent}} & \multirow{2}*{\SI{32 \pm 5}{\percent}} & 18.20 \\ 
 & & & & \mbox{[14.45, 21.95]} \\ \cmidrule(r){1-5}
 
 \multirow{2}*{Maximum} & \multirow{2}*{\SI{20 \pm 4}{\percent}} & \multirow{2}*{\SI{40 \pm 5}{\percent}} & \multirow{2}*{\SI{49 \pm 5}{\percent}} & 9.37 \\ 
 & & & & [7.52, 11.22] \\ 
 \bottomrule
\end{tabulary}
\label{tab:avg_max}
\end{table}

In Section \ref{sec:text_sim}, we discussed how the paper-reviewer SPECTER similarity is calculated between each paper $\Paper$ and reviewer $\Reviewer$ with two common methods of aggregation: max pooling and average pooling. 
So far, most of our analyses have focused on the setting where the colluding reviewer can curate their archives to contain a single paper, which is a scenario where max and average pooling similarity definitions become the same. 
In this section, we compare the two aggregation methods \emph{without} adversarial curation (i.e., all papers are retained) to investigate their differences when the similarity definitions are no longer the same. 

In the case where the colluding reviewer $\Reviewer$ is not allowed to make changes to their archive (that is, $\ArchiveAdv{} = \Archive{}$), we hypothesize that our attack would be more effective against the maximum aggregation method since abstract modifications can target just one paper in $\Archive{}$.  
We randomly sample without replacement 100 (paper $\Paper$, reviewer $\Reviewer$) pairs with natural rankings of 101 under each aggregation method. Then, for each pair, we run the automatic abstract modification attack procedure on the paper $\Paper$ to increase the similarity to reviewer $\Reviewer$ (no adversarial archive curation). We evaluate the attack success rates amongst the 100 samples for each scenario and enumerate our results in Table \ref{tab:avg_max}. We indeed find that our proposed attack is more successful under maximum aggregation. This result suggests that conferences that choose to use the more popular maximum aggregation method are generally \emph{more} susceptible to our proposed reviewer assignment attack.

\subsection{Correlation of Attack Success between NeurIPS 2022 and NeurIPS 2023} \label{sec:correlation}

\begin{figure}[b]
    \centering
    \includegraphics[scale=0.3]{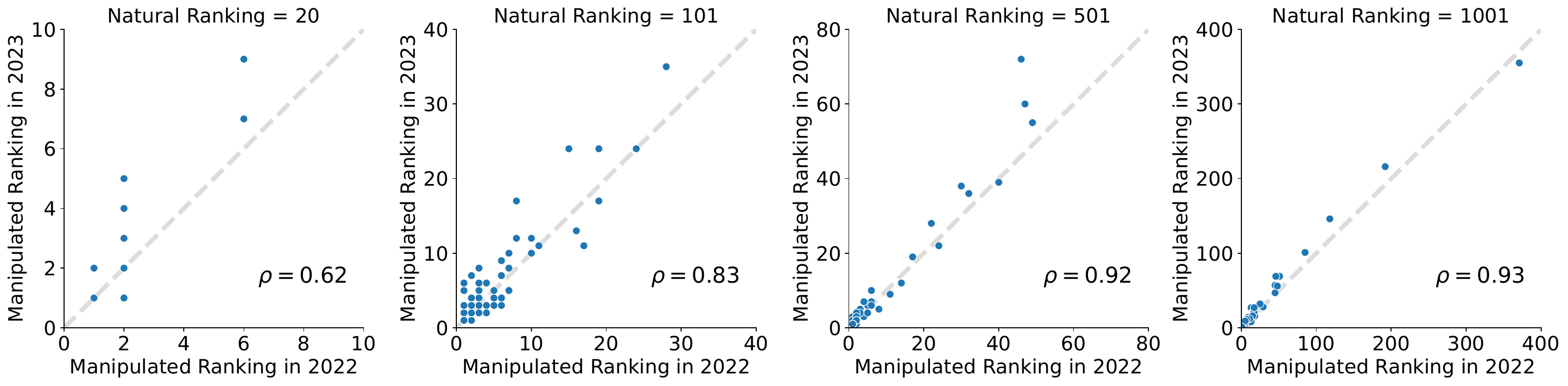}
    \caption{Manipulated rankings in 2022 and 2023 iterations of NeurIPS are strongly correlated, so colluders can estimate the manipulated rankings using previous year's data. Spearman's rank correlation coefficients $\rho$ is reported, and the dotted line is $y=x$.}
    \label{fig:correlation}
\end{figure}

An attacker can use publicly available data from the previous year's conference to train or validate their attack. For example, in Section \ref{sec:manual_result}, we discussed an effective early stopping heuristic that halts further modifications once the colluding reviewer becomes the top match amongst prior-year reviewers. The relationship between success of any attack in the previous year to the success in the target year is not clear a priori, 
and we investigate it in this section.

In Figure \ref{fig:correlation}, we calculate the manipulated rankings amongst both NeurIPS 2022 and NeurIPS 2023 reviewer pools for 300 colluding (paper $\Paper$, reviewer $\Reviewer$) pairs with natural rankings of 101 and 100 $\PaperReviewerPair$ pairs each for natural rankings of 20, 501, and 1001 from Section \ref{sec:automatic_result}.
As shown in Figure~\ref{fig:correlation}, we discover a strong correlation between the manipulated rankings in the 2022 (publicly available) and 2023 (unknown to attackers) iterations of NeurIPS. This implies that adversarial attackers can estimate attack success using previous year's data. 
The reviewer pools data many major ML/AI conferences publish give attackers a dataset to tune their modifications, knowing that being successful on the public data is often good enough.

\subsection{Human Subject Experiment on Detectability} \label{sec:human_subject_experiment}
In the previous sections, we found that our method can successfully increase the manipulated rankings between a paper and a colluding reviewer. However, the paper may also be assigned to honest reviewers, and in this section we describe a randomized control trial that we conduct to understand the perception of adversarial abstracts by unsuspecting human reviewers. The research question and study design were preregistered at \url{https://aspredicted.org/HXF_Z92}. This experiment was approved by an independent Institutional Review Board at Carnegie Mellon University (Federal Wide Assurance No: FWA00004206, IRB Registration No: IRB00000603). The broad {\bf research question} we looked to answer was: Are benign (control) and adversarial (experimental) abstracts the same to unsuspecting human reviewers? We also acknowledge that, as discussed below, this experimental design also had several limitations that can lead to an overestimation of the detectability of the adversarially modified abstract. 

\subsubsection{\textbf{Experiment design}} We recruited 59 participants with one opting out after debriefing (we initially employ deception to hide the true purpose of this study), making it a total of 58 participants. The participants are all either PhD students or those who have graduated with a PhD -- who have reviewed at ML/AI conferences in the past five years. Participants were recruited by emails and word of mouth at several U.S. universities, and each participant was compensated \$20. To simulate the behavior of an unsuspecting reviewer at an actual conference, we employ deception so that the participants have no knowledge of the adversarial manipulations prior to starting the study. 

We curated two pools of papers --- \textit{benign} and \textit{adversarial}, each containing modified versions of 914 real machine learning papers we collected from \href{https://arXiv.org}{arxiv.org}. To generate the adversarial version of each paper, we first select a reviewer that has natural ranking of 101 out of the NeurIPS 2023 reviewer pool, then we curate the adversarial archive $\ArchiveAdv{}$ (Section \ref{sec:reviewer}) and generate $\AbstractAdv$ by Algorithm \ref{alg:include_theme} and Algorithm \ref{alg:insert_keywords} under the \textit{fully automatic} mode for scalability. As a result of the adversarial attack, the colluding reviewer for each arXiv paper with an adversarial abstract had a manipulated ranking within the top-5 of all NeurIPS 2023 reviewers. Since the adversarial abstracts are LLM-generated without any human oversight, we believe that directly comparing them against the original human-written abstracts may introduce a confounder to our experiment that the adversarial abstracts are LLM-generated while the benign abstracts may be written by humans. To address this, in the benign (control) paper pool, we ask the LLM to paraphrase the original abstract. 

Each participant was assigned two personalized papers based on their expertise. Each paper assignment was chosen to be in either the \textit{benign} (control) or \textit{adversarial} (experimental) condition uniformly at random. 
Depending on whether the assignment is control or experimental, we computed the participant's SPECTER similarities to the pool of benign or adversarial papers, respectively. 
Using this process, we assigned each participant two distinct papers they are the most similar to.

\subsubsection{\textbf{Attack budgets and LLM prompts}} 
This experiment uses a different set of attack budgets and LLM prompts from the other experiments. The budgets and prompts in this experiment are tuned the same way as those in the rest of this paper, but they were erroneously tuned on the NeurIPS 2023 test data instead of the NeurIPS 2022 training data. However, we believe this should not affect the outcome of this experiment because the abstracts studied in this human subject experiment are being manipulated more than the automatically modified abstracts evaluated in earlier sections. In addition, we are not measuring the attack success rates, which would be affected by this error, but rather the differences (if any) between benign and adversarial abstracts to unsuspecting human reviewers. 

For the human study experiment, we chose $\Regenerations=3$, $\Alternations=2$ and $\Keywords=5$, and the prompts can be found in Appendix \ref{appx:hse_prompts}. Firstly, in the \IncludeRelatedWork{} operation, we prompt the LLM to follow the same rules and add only one sentence about the colluding reviewer's archive, just like in the rest of this paper. In the \InsertKeywords{} operation, we ask the LLM to add 12 keywords in this experiment, which is more than the 10 keywords ($\Regenerations=2$, $\Keywords=5$) we use in the rest of this paper. Finally, early stopping is also not used in the automatic abstracts modification process for this experiment. Due to the extra keywords added, the adversarially modified abstract tested in this experiment may be more detectable than it would be for the abstracts generated in previous sections. 

\subsubsection{\textbf{Experimental procedure}} After each participant signed up for our study, we emailed them two PDFs of the papers they are assigned to review. We asked participants to notify us if they had seen either paper prior to this study, and we assigned them new papers if they had seen them before. These PDFs are rendered from the \LaTeX{} source available on arXiv, except we replaced the original abstracts with the adversarial or benign versions of the abstract. In addition, we anonymized the arXiv papers by removing the author names. After they completed the tasks, we debriefed the participants since deception was used and gave them the option to withdraw from the study after learning about the real purpose of this study.

\subsubsection{\textbf{Survey}} \label{sec:exp_procedure}
Ideally, we would like to ask participants to write full reviews of the papers, but such time commitment was not feasible since each review can take hours. Therefore, we asked the participants to take ten minutes to skim and complete a `mini-review' for each paper. Each mini-review consists of the following questions: 

\begin{enumerate}[label=\alph*.]
    \item Would you be able to review this paper given your expertise? [Yes/No]
    \item Is the abstract of the paper consistent with the contents in the paper? [Yes/No]
    \item Does the abstract of the paper seem coherent? [Yes/No]
    \item If you answered ``No'' to any of the questions, please explain. [Text box]
\end{enumerate}

\subsubsection{\textbf{Limitations}}
To make the human subject experiment scalable and participant recruitment practical, we made several design choices that also manifest as limitations. Each of these limitations can overestimate the identifiability of the adversarial abstracts.
\begin{itemize}
    \item We generated the adversarial abstracts automatically without any degree of human oversight for scalability. This may not reflect operation of actual colluders in the real world, where malicious authors can at least look over the manipulated abstract to check for detectability. 
    \item Due to the scale, we were not able to edit the body of the papers sent to participants for review. However, since having topics and keywords in the abstract that are not in the rest of the paper can be suspicious, malicious authors in practice could also edit the body of the paper (e.g., introduction or related work) to contain those words. 
    \item To make recruitment practical, we asked participants to write mini-reviews that are focused on the abstracts. This may lead to participants reading the abstracts much more carefully than usual, potentially cross-referencing the abstract and the rest of the paper multiple times. 
    \item In practice, the authors separately enter their (title and) abstract in text boxes on a web interface, which are used to  compute the similarities with reviewers. This entered abstract may potentially have differences with the abstract in the paper's PDFs, and it is the PDFs that are generally read more carefully by reviewers.
\end{itemize}

Finally, we note that the monetary compensation may have induced some participants to pay less attention than others (e.g., if their participation was solely for compensation).

\subsubsection{\textbf{Results}} We collected a total of 116 mini-reviews from participants, comprising 49 reviews for the control group and 67 reviews for the experimental group. Out of 116 mini-reviews, 51 of the reviews included free response comments for Question 4 in the survey (Section \ref{sec:exp_procedure}). As we discuss below, the results are mixed. First, we consider the total counts of Yes/No responses about the paper-reviewer expertise alignment as well as the consistency and coherence of the abstracts. Table \ref{tab:study_stats} summarizes the findings, where we see that control abstracts are considered significantly more consistent than experimental abstracts, but no significant difference is found in expertise alignment and abstract coherence.

\begin{table}[b]
\caption{Fisher's Two-Sided Exact Test results and participants' ``No'' response proportions. The p-values are adjusted using Benjamini-Hochberg correction \cite{benjamini1995controlling} for multiple testing.}
\centering
\tymax=300pt
\begin{tabulary}{\linewidth}{@{}CCCCC@{}} \toprule
    Evaluation Category & p-value & $1-\beta$ (Power) & ``No'' Rate (Experimental) & ``No'' Rate (Control) \\ \midrule
    Expertise & 0.65 & 0.06 & \SI{22.7}{\percent} & \SI{18.8}{\percent} \\ 
    Consistency & 0.03 & 0.74 & \SI{24.2}{\percent} & \SI{6.3}{\percent} \\
    Coherence & 0.255 & 0.27 & \SI{25.8}{\percent} & \SI{14.6}{\percent} \\ \bottomrule
\end{tabulary}
\label{tab:study_stats}
\end{table}

\begin{table}[tb]
\caption{Complaint types from comments in the collected reviews.}
\centering
\tymax=300pt
\begin{tabulary}{\linewidth}{@{}JCC@{}} \toprule
    Type of complaint & Control & Experimental \\  \midrule
    Issues with the writing style & \SI{8.2}{\percent} & \SI{25.4}{\percent}  \\
    Abrupt transitions \& poor organization & \SI{2.0}{\percent} & \SI{4.5}{\percent}\\ 
    Nonsensical or incorrect claims & \SI{4.1}{\percent} & \SI{10.4}{\percent}   \\
    Contains things never mentioned in the paper & \SI{4.1}{\percent} & \SI{13.6}{\percent}  \\
    Not representative of the paper content & \SI{2.0}{\percent} & \SI{4.5}{\percent}  \\
    \hline
    Irregularities related to \IncludeRelatedWork{} & -- & 6\% \\
    Irregularities related to \InsertKeywords{} & -- & 16\% \\
    Not related to either & -- & 82\% \\
    \bottomrule
\end{tabulary}
\label{tab:comments}
\end{table}
In Table~\ref{tab:comments} we analyze the free response comments provided. Each review that indicated either poor coherence or inconsistency includes comments, and all comments are between 7 and 189 words.
We identify five main types of complaints in the comments, and we report the percentages of reviews in control and experimental groups that mention each type of complaint. The percentages are out of all reviews collected for each group (49 control reviews and 67 experimental reviews), and there may be multiple types of complaints mentioned in each comment. The most prevalent issue for both groups is writing style, which is related to word choice, tone, and readability. Surprisingly, complaints like ``contains things never mentioned in the paper'' exist for both benign and adversarial abstracts. 
To investigate whether these complaints are due to LLM hallucinations during the benign paraphrasing of control abstracts rather than the original contents uploaded to arXiv, we manually examined the control and original abstracts related to each complaint. 
We find that all types of complaints have at least one paper which we believe the complaint applies to the original abstract. 

Lastly, we investigated whether the review comments for the adversarial abstracts mentioned issues that are directly related to the modification operations proposed in this work. In the experimental group, we noticed that 6\% of the responses are related to \IncludeRelatedWork{}, 16\% of the reponses are related to \InsertKeywords{}, and 82\% of the responses are not directly related to either operations. Some responses may be related to both operations, so the percentages do not sum to 100\%. Finally, none of the participants identified malicious intent in any of the abstracts. Although it is not surprising that no participant identified malicious intent given the use of deception, we believe this finding remains significant, because real-world peer reviewers likewise are not explicitly instructed to look for signs of collusion.

In summary, we find that \emph{no participants suspected malicious intent}, but we identify higher rates of complaints about the coherence and consistency of adversarial abstracts when compared to benign abstracts. Problems in coherence and consistency may have benign causes (e.g., negligence, bad writing, or authors not having English as their first language), giving colluding authors plausible deniability if accused of malicious manipulations. With the proliferation of LLM-edited abstracts, the (in)ability to distinguish between adversarial and benign abstracts is even more dire because the complaint rates of abstracts with adversarial manipulations are not much higher than benign manipulations with an LLM. Furthermore, the attack we evaluate is fully automatic; the human-in-the-loop version is likely to raise less complaints because the attacker can manually catch obvious inconsistencies and 
iterate, as we explain in our method (Section \ref{sec:author}).

\section{Discussion}
In this work, we identify the vulnerability of current text-based reviewer-assignment systems in AI/ML venues to manipulation by collusion rings. This contradicts common perceptions that text-based automatic reviewer matching is resistant to such tactics. Our work suggests that a simple and practical attack procedure can effectively manipulate paper-reviewer similarities and hence manipulate automated reviewer assignment at many ML/AI conferences. Our attacks have a high success rate in matching to a targeted colluding reviewer. In the human subject experiment testing for attack detectability, no participant detected malicious intent for the manipulated abstracts. While there were complaints about coherence and consistency, benign abstracts edited by LLMs elicited similar complaints, suggesting plausible deniability. 

This work raises awareness about the vulnerabilities of the current reviewer matching systems, and we are helping to address the these issues by informing the concerned conference management platform. To guard against attacks such as those described in this paper, platform and algorithm changes based on our findings and recommendations have been implemented platform-wide and used at a major AI/ML conference. 
In total, we underscore the need for enhanced robustness in reviewer assignment algorithms to protect the integrity of the peer review process.

\subsection{Recommendations for mitigation}

To improve the robustness of text-based reviewer matching systems against adversarial attacks, we offer several recommendations based on our findings:

\begin{itemize}
	\item \textbf{Increase reviewer profile requirements:} Our results indicate that requiring reviewers to have more extensive publication archives reduces the effectiveness of manipulation. This makes it more difficult for colluding reviewers to align their profiles strategically with targeted abstracts.
	
	\item \textbf{Pooling for similarity calculations:}  To combine the similarity scores between a submitted paper and a reviewer's multiple papers, using average pooling instead of max pooling can help reduce the impact of targeted manipulations, since average pooling lessens the influence of any single manipulated abstract on the overall similarity. However, if max pooling results in more accurate (honest) matches and is generally preferred, a compromise could be to use an order statistic such as the 75th percentile of the similarity scores across the reviewer's papers. This approach balances robustness to manipulation with preserving high-quality matches.
	
	\item \textbf{Increase reviewer awareness:} Educating reviewers about the possibility of manipulations can prompt them to be more vigilant when evaluating abstracts and papers.
	
	\item \textbf{Introduce randomness in assignments:} No component of automated assignment methods is entirely immune to manipulation. This highlights the importance of adopting a broader mitigation strategy such as introducing a degree of randomness into the assignment process~\cite{jecmen2020mitigating}. Such randomness provably prevents adversaries from reliably influencing assignments.

    \item \textbf{Consider robustness of similarity scores:} Developers of similarity algorithms should pay attention to impart some robustness so the similarity scores are less susceptible to adversarial manipulations to the abstract.
\end{itemize}
It is important to recognize that these mitigation strategies may involve a trade-off between robustness and the quality of the match. Therefore, investigating the extent of this trade-off and establishing optimal points along it is valuable, allowing program chairs to decide where to operate on this spectrum. As a positive example, for the randomized assignments proposed in~\cite{jecmen2020mitigating}, both laboratory evaluations and multiple real world deployments have found that the robustness imparted by such randomization comes at very little cost to the optimality of the match.

\subsection{Limitations}
While our study provides valuable insights, it has several limitations that offer avenues for future research. Our analysis considers scenarios where authors collude with a single reviewer within a panel of 3-6 reviewers. Expanding this to cases where multiple reviewers colluding with the author simultaneously could provide a more comprehensive understanding of the system's vulnerabilities. We also did not factor in common constraints such as reviewer and paper load limits, conflicts of interest, or geographical considerations. Incorporating these constraints could affect the attack's efficacy and the generalizability of some of our findings. Additionally, although the \AutoMode{} mode enables large-scale experimentation, the abstract modifications may not fully reflect the operations of real colluders due to the absence of human supervision.
Finally, it would be useful to evaluate the attack success rates in other venues. 

In our human subject experiment, the adversarially modified abstract tested may be more detectable than it would be for the adversarial abstracts evaluated in all other experiments, due to the higher attack budget that was erroneously tuned for the human subject experiment. In addition, in a laboratory experiment like ours, we could not fully replicate the complexities of real-world reviewer behavior. Factors such as varying levels of expertise, bias, and attention could influence outcomes in practical settings. We also do not consider how different reviewers could influence each other. 

\section*{Acknowledgments} 
This work was supported in part by grants ONR N000142212181 and NSF 2200410, 1942124, and a J.P. Morgan research scholar program. AR gratefully acknowledges support from the AI2050 program at Schmidt Sciences (Grant \#G2264481), Google Research Scholar program, Apple, Cisco, Open Philanthropy, and NSF 2310758.

\section*{Ethical considerations}
The key stakeholders of this research are ML/AI conference venues, submission authors, and its research community. As other Computer Science (CS) fields grow in size, our findings may be more widely applicable. The main risk of this work is the \textit{disclosure} of a serious vulnerability in reviewer matching systems, which could allow unfair advantages and harm publication quality. To mitigate this risk, we had informed the concerned conference management platform before submission and worked with them to address these vulnerabilities using the defenses we propose. \textbf{These safeguards are now in place and are being used.}

\textit{Collusions and attacks described in this paper may already be happening in practice}, yet most researchers may not be aware of these vulnerabilities. To promote transparency, we make our attack algorithm and adversarial abstract examples publicly available. 
By raising awareness of these vulnerabilities, we aim to empower non-colluding reviewers to be more vigilant.
Furthermore, attacks described in this work are difficult to detect and often come with \textit{plausible deniability}. Given the challenge of proving intent based on abstract content or profile changes, proactive mitigation is crucial rather than relying solely on post hoc detection and enforcement. 

Finally, while this paper finds that colluders can manipulate text similarity to get high rankings with high success, it is important to note that the results do not imply any high ranking reviewer-paper pair in past data are actually colluders. Any successful attack pertains to the abstract and/or reviewer profile modified in this experiment and not to the original abstract or profile in the conference. Furthermore, the success rates are at an aggregate level, and nothing in the paper (or in any of its artifacts) should be explicitly construed to suggest collusion or a reviewing relationship.

\paragraph{Human Subject Experiments}  
This experiment was approved by an independent
Institutional Review Board at Carnegie Mellon University
(Federal Wide Assurance No: FWA00004206, IRB Registration No: IRB00000603). To simulate real reviewer behavior, we used \textit{deception} to conceal the study’s purpose, as disclosure is likely to cause participants to act differently and be hyper-aware. 
All participants were debriefed and informed why deception was necessary. Participants could opt out at any time, including after debriefing, and only one participant opted-out (after debriefing).
To ensure participant privacy, we only share anonymized, aggregated statistics of the collected data.

\paragraph{Datasets}
We considered the risks of releasing reviewer and author names from the NeurIPS 2022 and 2023 datasets, given the adversarial context, but believe they are negligible. 
Again, we make no claims that can be construed to suggest collusion or reviewing relationships.
Furthermore, all data we release is already public, making the added risk low. 

\paragraph{Abstracts}  
We clearly label all modified abstracts to avoid damages to the original paper and authors. Since the original abstracts are available on OpenReview or arXiv, the risk of reputational harm is minimal.
Finally, the Creative Commons Public Domain Dedication (CC0 1.0) licenses of paper metadata on OpenReview and arXiv allow us to modify and distribute the abstracts (license links: \href{https://openreview.net/legal/terms}{OpenReview}, \href{https://info.arxiv.org/help/license/index.html}{arXiv}).

\paragraph{} All in all, we believe the benefits of this research outweigh the risks. 
We hope our findings and recommendations can help the research community be more vigilant and robust to such malicious activities in scientific peer review.

\section*{Open Science}
Our research artifacts listed below are available on Zenodo at \url{https://doi.org/10.5281/zenodo.15588237}. 
\paragraph{Adversarial Examples}
All adversarial abstracts evaluated in the results section are released. This includes 25 \HumanMode{} adversarial abstracts from Section~\ref{sec:manual_result}; 600 \AutoMode{} adversarial abstracts from Section~\ref{sec:automatic_result}, with 300 for natural ranking of 101 and 100 each for 20, 501, and 1001; 400 \AutoMode{} adversarial abstracts from Section~\ref{sec:arclen}, with 100 each for reviewer archive lengths of 1, 2, 5, and 10; and 200 \AutoMode{} adversarial abstracts from Section~\ref{sec:pooling}, with 100 each for average and maximum aggregation methods. The adversarial abstracts used in Section~\ref{sec:correlation} are the same as those in Section~\ref{sec:automatic_result}. Additionally, all 914 \AutoMode{} adversarial abstracts corresponding to the arXiv papers used in the human subject experiment (Section~\ref{sec:human_subject_experiment}) are released.
\paragraph{Datasets}
We release both the NeurIPS 2023 dataset and the NeurIPS 2022 dataset described in Section \ref{sec:exp_setup}. We also release the dataset of 914 arXiv papers and their control abstracts (from benign LLM paraphrasing) used in Section \ref{sec:human_subject_experiment}.
\paragraph{Code}
We release our code to execute the attacks and the notebooks/scripts used for evaluating the results in this paper. Scripts used to sample paper-reviewer pairs for evaluation (see ``Evaluation samples'' in Section \ref{sec:exp_setup}) are also provided.
\paragraph{LLM Prompts}

We release a total of six LLM prompts: five prompts corresponding to the \IncludeRelatedWork{} and \InsertKeywords{} operations across \HumanMode{}, \AutoMode{}, and the human study, and one prompt for paraphrasing control abstracts for the human study. Additionally, we provide four few-shot examples used in our experiments, with two examples per modification operation.

\paragraph{Human Subject Data}
Since study participants are assigned personalized papers to review based on their past publications, the raw data of reviews, even if anonymized, can still compromise their privacy. 
According to IRB requirements and our promise to study participants, we only release the anonymized and aggregated statistics as presented in Section~\ref{sec:human_subject_experiment}.

\bibliographystyle{alpha}
\bibliography{main}

\newcommand{\etalchar}[1]{$^{#1}$}
\begin{thebibliography}{WGNBK19}

\bibitem[ACDK19]{ailamaki2019sigmod}
Anastasia Ailamaki, Periklis Chrysogelos, Amol Deshpande, and Tim Kraska.
\newblock The sigmod 2019 research track reviewing system.
\newblock {\em ACM SIGMOD Record}, 48(2):47--54, 2019.

\bibitem[BBN22]{boehmer2022combating}
Niclas Boehmer, Robert Bredereck, and Andr{\'e} Nichterlein.
\newblock Combating collusion rings is hard but possible.
\newblock In {\em Proceedings of the AAAI Conference on Artificial Intelligence}, volume~36, pages 4843--4850, 2022.

\bibitem[BH95]{benjamini1995controlling}
Yoav Benjamini and Yosef Hochberg.
\newblock Controlling the false discovery rate: a practical and powerful approach to multiple testing.
\newblock {\em Journal of the Royal statistical society: series B (Methodological)}, 57(1):289--300, 1995.

\bibitem[CFB{\etalchar{+}}20]{cohan2020specter}
Arman Cohan, Sergey Feldman, Iz~Beltagy, Doug Downey, and Daniel~S Weld.
\newblock {SPECTER}: Document-level representation learning using citation-informed transformers.
\newblock {\em arXiv preprint arXiv:2004.07180}, 2020.

\bibitem[CZ13a]{charlin13tpms}
L.~Charlin and R.~S. Zemel.
\newblock The {T}oronto {P}aper {M}atching {S}ystem: An automated paper-reviewer assignment system.
\newblock In {\em ICML Workshop on Peer Reviewing and Publishing Models}, 2013.

\bibitem[CZ13b]{charlin2013toronto}
Laurent Charlin and Richard Zemel.
\newblock The toronto paper matching system: an automated paper-reviewer assignment system.
\newblock {\em The International Conference on Learning Representations}, 2013.

\bibitem[EQM{\etalchar{+}}23]{eisenhofer2023no}
Thorsten Eisenhofer, Erwin Quiring, Jonas M{\"o}ller, Doreen Riepel, Thorsten Holz, and Konrad Rieck.
\newblock No more reviewer\# 2: Subverting automatic paper-reviewer assignment using adversarial learning.
\newblock In {\em 32nd USENIX Security Symposium (USENIX Security 23)}, pages 5109--5126, 2023.

\bibitem[ERLD17]{ebrahimi2017hotflip}
Javid Ebrahimi, Anyi Rao, Daniel Lowd, and Dejing Dou.
\newblock {HotFlip}: White-box adversarial examples for text classification.
\newblock {\em arXiv preprint arXiv:1712.06751}, 2017.

\bibitem[GWC{\etalchar{+}}18]{guo2018k}
Longhua Guo, Jie Wu, Wei Chang, Jun Wu, and Jianhua Li.
\newblock {K-loop} free assignment in conference review systems.
\newblock In {\em 2018 International Conference on Computing, Networking and Communications (ICNC)}, pages 542--547. IEEE, 2018.

\bibitem[HSC21]{chang2021mfr}
Andrew~McCallum Haw-Shiuan~Chang.
\newblock Knuth: Computers and typesetting, 2021.
\newblock \url{https://www.overleaf.com/project/5f359923225f06000134ea95}. Last accessed 14 April 2024.

\bibitem[JJZS20]{jin2020bert}
Di~Jin, Zhijing Jin, Joey~Tianyi Zhou, and Peter Szolovits.
\newblock Is {BERT} really robust? a strong baseline for natural language attack on text classification and entailment.
\newblock In {\em Proceedings of the AAAI conference on artificial intelligence}, volume~34, pages 8018--8025, 2020.

\bibitem[JSFA24]{jecmen2024detection}
Steven Jecmen, Nihar~B Shah, Fei Fang, and Leman Akoglu.
\newblock On the detection of reviewer-author collusion rings from paper bidding.
\newblock {\em arXiv preprint arXiv:2402.07860}, 2024.

\bibitem[JSFC22]{jecmen2022tradeoffs}
Steven Jecmen, Nihar~B Shah, Fei Fang, and Vincent Conitzer.
\newblock Tradeoffs in preventing manipulation in paper bidding for reviewer assignment.
\newblock In {\em ICLR workshop on {ML} Evaluation Standards}, 2022.

\bibitem[JYC{\etalchar{+}}23]{jecmen2022dataset}
Steven Jecmen, Minji Yoon, Vincent Conitzer, Nihar~B. Shah, and Fei Fang.
\newblock A dataset on malicious paper bidding in peer review.
\newblock In {\em TheWebConf}, 2023.

\bibitem[JZL{\etalchar{+}}20]{jecmen2020mitigating}
Steven Jecmen, Hanrui Zhang, Ryan Liu, Nihar Shah, Vincent Conitzer, and Fei Fang.
\newblock Mitigating manipulation in peer review via randomized reviewer assignments.
\newblock {\em Advances in Neural Information Processing Systems}, 33:12533--12545, 2020.

\bibitem[KSM19]{kobren19localfairness}
Ari Kobren, Barna Saha, and Andrew McCallum.
\newblock Paper matching with local fairness constraints.
\newblock In {\em ACM KDD}, 2019.

\bibitem[LBNZ{\etalchar{+}}24]{leyton2024matching}
Kevin Leyton-Brown, Yatin Nandwani, Hedayat Zarkoob, Chris Cameron, Neil Newman, and Dinesh Raghu.
\newblock Matching papers and reviewers at large conferences.
\newblock {\em Artificial Intelligence}, 331:104119, 2024.

\bibitem[Lit21]{littman2021collusion}
Michael~L Littman.
\newblock Collusion rings threaten the integrity of computer science research.
\newblock {\em Communications of the ACM}, 64(6):43--44, 2021.

\bibitem[LMG{\etalchar{+}}20]{li2020bert}
Linyang Li, Ruotian Ma, Qipeng Guo, Xiangyang Xue, and Xipeng Qiu.
\newblock Bert-attack: Adversarial attack against bert using bert.
\newblock {\em arXiv preprint arXiv:2004.09984}, 2020.

\bibitem[MJMZ23]{mysore2023editable}
Sheshera Mysore, Mahmood Jasim, Andrew McCallum, and Hamed Zamani.
\newblock Editable user profiles for controllable text recommendation.
\newblock {\em arXiv preprint arXiv:2304.04250}, 2023.

\bibitem[MM07]{mimno07topicbased}
David Mimno and Andrew McCallum.
\newblock Expertise modeling for matching papers with reviewers.
\newblock In {\em KDD}, 2007.

\bibitem[MSLL17]{markwood2018mirage}
Ian Markwood, Dakun Shen, Yao Liu, and Zhuo Lu.
\newblock {PDF} mirage: Content masking attack against {Information-Based} online services.
\newblock In {\em 26th USENIX Security Symposium (USENIX Security 17)}, pages 833--847, Vancouver, BC, August 2017. USENIX Association.

\bibitem[ORA{\etalchar{+}}22]{ostendorff2022neighborhood}
M~Ostendorff, N~Rethmeier, I~Augenstein, et~al.
\newblock Neighborhood contrastive learning for scientific document representations with citation embeddings.
\newblock {\em arXiv preprint arXiv:2202.06671}, 2022.

\bibitem[PC24]{ICLR2024chairs}
{ICLR}~2024 {P}rogram {C}hairs.
\newblock Slide from {ICLR} 2024 program chair presentation, 2024.
\newblock \url{https://twitter.com/chriswolfvision/status/1787886748434878769}. Last accessed 16 May 2024.

\bibitem[PZ22]{payan2021will}
Justin Payan and Yair Zick.
\newblock I will have order! optimizing orders for fair reviewer assignment.
\newblock In {\em Proceedings of the 21st International Conference on Autonomous Agents and Multiagent Systems}, pages 1711--1713, 2022.

\bibitem[SDC{\etalchar{+}}22]{singh2022scirepeval}
Amanpreet Singh, Mike D'Arcy, Arman Cohan, Doug Downey, and Sergey Feldman.
\newblock Scirepeval: A multi-format benchmark for scientific document representations.
\newblock {\em arXiv preprint arXiv:2211.13308}, 2022.

\bibitem[Sha22]{shah2022challenges}
Nihar~B Shah.
\newblock Challenges, experiments, and computational solutions in peer review.
\newblock Communications of the ACM. Preprint available at \url{https://www.cs.cmu.edu/~nihars/preprints/SurveyPeerReview.pdf}, June 2022.

\bibitem[SSS21]{stelmakh2018forall}
Ivan Stelmakh, Nihar Shah, and Aarti Singh.
\newblock {PeerReview4All}: Fair and accurate reviewer assignment in peer review.
\newblock {\em Journal of Machine Learning Research}, 2021.

\bibitem[TJ19]{tran2019pdfphantom}
Dat Tran and Chetan Jaiswal.
\newblock {PDFPhantom}: Exploiting pdf attacks against academic conferences' paper submission process with counterattack.
\newblock In {\em 2019 IEEE 10th Annual Ubiquitous Computing, Electronics \& Mobile Communication Conference (UEMCON)}, pages 0736--0743, 2019.

\bibitem[Vij20a]{vijaykumar2020ieee}
T.~N. Vijaykumar.
\newblock Potential organized fraud in {ACM}/{IEEE} computer architecture conferences, 2020.
\newblock \url{https://medium.com/@tnvijayk/potential-organized-fraud-in-acm-ieee-computer-architecture-conferences-ccd61169370d}. Last accessed 29 April 2024.

\bibitem[Vij20b]{vijaykumar2020asplos}
T.~N. Vijaykumar.
\newblock Potential organized fraud in on-going {ASPLOS} reviews, 2020.
\newblock \url{https://medium.com/@tnvijayk/potential-organized-fraud-in-on-going-asplos-reviews-874ce14a3ebe}. Last accessed 29 April 2024.

\bibitem[WGNBK19]{wieting2019simple}
John Wieting, Kevin Gimpel, Graham Neubig, and Taylor Berg-Kirkpatrick.
\newblock Simple and effective paraphrastic similarity from parallel translations.
\newblock In {\em ACL}, pages 4602--4608, Florence, Italy, July 2019.

\bibitem[WGW{\etalchar{+}}21]{wu2021making}
Ruihan Wu, Chuan Guo, Felix Wu, Rahul Kidambi, Laurens Van Der~Maaten, and Kilian Weinberger.
\newblock Making paper reviewing robust to bid manipulation attacks.
\newblock In {\em International Conference on Machine Learning}, pages 11240--11250. PMLR, 2021.

\bibitem[XJSF24]{xu2024one}
Yixuan Xu, Steven Jecmen, Zimeng Song, and Fei Fang.
\newblock A one-size-fits-all approach to improving randomness in paper assignment.
\newblock {\em Advances in Neural Information Processing Systems}, 36, 2024.

\end{thebibliography}

~\\
\noindent\appendices

\noindent{\bf \Large Appendices}

\section{Adversarial Abstract Modification Algorithms} \label{appx:algorithms}

Before introducing the formal algorithms of the two abstract modification operations, we define a few useful helper functions: 
\begin{enumerate}
    \item \ConstraintsCheck{$\AbstractAdv$} returns true if abstract $\AbstractAdv$ is coherent and consistent.
    
    \item \SimilarityCheck{$\Title$}{$\AbstractAdvEdit$}{$\AbstractAdv$}{$\ArchiveAdv{}$}{$\SimDelta$} queries the SPECTER model and returns true if the similarity of abstract $\AbstractAdvEdit$ is higher than (or at least comparable to) the similarity of abstract $\AbstractAdv$, that is, $\SimilarityAbstract{\Title}{\AbstractAdvEdit}{\ArchiveAdv{}} + \SimDelta > \SimilarityAbstract{\Title}{\AbstractAdv}{\ArchiveAdv{}}$. The $\delta$ parameter represents a small non-negative value, for cases when the new edits added in $\AbstractAdvEdit$ do not have to be more similar to reviewer $\Reviewer$ than the $\AbstractAdv$. 
    
    \item  \EarlyStoppingCheck{$\AbstractAdv$}{$\ArchiveAdv{}$} returns true if the colluding reviewer $\Reviewer$ (with the curated adversarial archive $\ArchiveAdv{}$) is the most-similar reviewer for the paper $\Paper$ (with the adversarial abstract $\AbstractAdv$) amongst some (potentially proxy) set of reviewers. In the proposed method, abstracts are modified in a multistep process, so an \textit{early stopping check} may be desirable to stop further modifications if the attack is envisaged to be successful. The use of early stopping is \textbf{optional} in our algorithm, since it is designed to trade off between attack effectiveness and abstract modification strength. In one of our experiments, we use the following early stopping heuristic: \emph{if the colluding reviewer is the most-similar reviewer for the colluding paper among all reviewers in the previous edition of the conference, no further abstract modifications are made}. 
    This heuristic is feasible and realistic because many major AI/ML conferences publish their reviewer pools from previous years. 
    We validate the effectiveness of our early stopping heuristic in Section \ref{sec:manual_result} and further investigate the use of proxy sets in Section \ref{sec:correlation}. 
\end{enumerate}

\subsection{The \IncludeRelatedWork{} Algorithm}\label{appx:include_related_work_alg}

Algorithm~\ref{alg:include_theme} formalizes the \IncludeRelatedWork{} operation, and it shows both \HumanMode{} and \AutoMode{} modes.
In \textit{human-in-the-loop} mode, authors can incrementally edit abstracts until coherent and consistent while maintianing high similarity. We have $\SimDelta$, which is a small positive value that allows such incrementally edited $\AbstractAdvEdit$ to trade slightly lower similarity than the current $\AbstractAdv$ for ensuring unsuspicious adversarial abstracts. We subjectively pick $\SimDelta$ to be a small value (around $0.01$) without systematic tuning.

\begin{algorithm}[htbp]  
    \caption{\IncludeRelatedWork{} Operation}\label{alg:include_theme}
    \hspace*{\algorithmicindent} \textbf{Input:} \parbox[t]{.8\linewidth}{ paper title $\Title$, paper abstract $\Abstract$, and adversarial reviewer archive $\ArchiveAdv{\Reviewer}$}\\
    \hspace*{\algorithmicindent} \textbf{Output:} \parbox[t]{.8\linewidth}{$\AbstractAdv$, an adversarial abstract that increases similarity to reviewer $\Reviewer$ by adding themes from $\ArchiveAdv{\Reviewer}$.}
    \vspace{0.1cm}
    \begin{algorithmic}[1]
    \Function{\IncludeRelatedWorkFunctionName{}}{$\Title$, $\Abstract$, $\ArchiveAdv{\Reviewer}$}
    \State $\AbstractAdvVersion{0} \gets \Abstract$ 
    \For{$i=1,\ldots,\Regenerations$}
        \If{Early stopping is used \textbf{and} \EarlyStoppingCheck{$\AbstractAdvVersion{i-1}$}{$\ArchiveAdv{\Reviewer}$}
        } 
            \State \textbf{return} $\AbstractAdvVersion{i-1}$
        \EndIf
        \If{mode == \HumanMode{}}
            \State $\AbstractAdv \gets$ \parbox[t]{0.8\linewidth}{A modified abstract including the main themes from $\ArchiveAdv{\Reviewer}$ into the original abstract $\Abstract$. The human author can modify the abstract to add themes from $\ArchiveAdv{\Reviewer}$ or add manual edits to the an LLM-generated version (see prompt in Appendix \ref{appx:include_themes_manual_prompt}).}
            \Repeat \Comment{Make incremental edits to $\AbstractAdv$.}
                \State $\AbstractAdvEdit \gets$ A draft with manual edits on $\AbstractAdv$ towards being consistent and coherent.
                \If{\SimilarityCheck{$\Title$}{$\AbstractAdvEdit$}{$\AbstractAdv$}{$\ArchiveAdv{\Reviewer}$}{$\SimDelta$} is true
                }
                    \State $\AbstractAdv \gets \AbstractAdvEdit$ \Comment{Update $\AbstractAdv$ if the similarity after edits do not drop dramatically.} 
                \EndIf
            \Until{\ConstraintsCheck{$\AbstractAdv$} is true}
        \Else \Comment{\AutoMode{} mode}
            \State $\AbstractAdv \gets$ \parbox[t]{0.8\linewidth}{An LLM generated adversarial abstract that includes themes from $\ArchiveAdv{\Reviewer}$ into the original abstract $\Abstract$. To reduce nonsensical abstract generations, we prompt the LLM to follow a format -- the generated abstract should say it takes inspiration from ideas in $\ArchiveAdv{\Reviewer}$ in one sentence (Appendix \ref{appx:include_themes_auto_prompt}).}
        \EndIf
        \vspace{0.1cm}
        \State $\AbstractAdvVersion{i} \gets \AbstractAdv$ 
    \EndFor
    \State \textbf{return} 
    $\AbstractAdvVersion{j} \in \argmax_{i \in [\Regenerations]}  \SimilarityAbstract{\Title}{\AbstractAdvVersion{i}}{\ArchiveAdv{\Reviewer}}$ of all $i=0,\ldots,\Regenerations$
    \EndFunction
    \end{algorithmic}
\end{algorithm}

\newpage
\subsection{The \InsertKeywords{} Algorithm}\label{appx:insert_keywords_alg}

Algorithm~\ref{alg:insert_keywords} formalizes the \InsertKeywords{} operation, and it also shows both \HumanMode{} and \AutoMode{} modes. We present \FindKeywords{} subroutine separately in the next part (Appendix \ref{appx:find_keywords}). 

\begin{algorithm}[h]  
    \caption{\InsertKeywords{} Operation}\label{alg:insert_keywords}
    \hspace*{\algorithmicindent} \textbf{Input:} \parbox[t]{.85\linewidth}{$\Title$ (the title of $\Paper$), $\AbstractAdv$ (the adversarial abstract from \IncludeRelatedWork{}), and $\ArchiveAdv{}$ (the adversarial archive of reviewer $\Reviewer$)}\\
    \hspace*{\algorithmicindent} \textbf{Output:} \parbox[t]{.85\linewidth}{$\AbstractAdv$, an adversarial abstract that increases similarity to $\Reviewer$ by adding keywords from $\ArchiveAdv{\Reviewer}$.}
    \vspace{0.1cm}
    \begin{algorithmic}[1]
    \Function{\InsertKeywordsFunctionName{}}{$\Title$, $\AbstractAdv$, $\ArchiveAdv{}$}
    \State $\AbstractAdvVersion{0} \gets \AbstractAdv$
    \For{$i=1,\ldots,\Alternations$}
        \If{Early stopping is used \textbf{and} \EarlyStoppingCheck{$\AbstractAdvVersion{i-1}$}{$\ArchiveAdv{}$}} 
            \State \textbf{return} $\AbstractAdvVersion{i-1}$
        \EndIf
        \State \KeywordsList{} $\gets$ \FindKeywords{($\Title$, $\AbstractAdv$, $\ArchiveAdv{}$, $\Keywords$)} \Comment{Algorithm \ref{alg:keywords}}
        \If{mode == \HumanMode{}}
            \For{each word $\word{}$ in \KeywordsList{}}
                \Repeat
                    \State $\AbstractAdvEdit \gets$ A new draft with one way $\word{}$ can be inserted into $\AbstractAdv$.
                    \If{\ConstraintsCheck{$\AbstractAdvEdit$} is true \textbf{and} \SimilarityCheck{$\Title$}{$\AbstractAdvEdit$}{$\AbstractAdv$}{$\ArchiveAdv{}$}{$0$} is true}
                        \State $\AbstractAdv \gets \AbstractAdvEdit$
                    \EndIf
                \Until{Up to five drafts $\AbstractAdvEdit$ have been generated for inserting $\word{}$}

            \EndFor
        \Else \Comment{\AutoMode{} mode.}
            \State $\AbstractAdv \gets$ \parbox[t]{0.8\linewidth}{Adversarial abstract generated by LLM to incorporate all \textit{keywords} into the current $\AbstractAdv$. The LLM is prompted to leave out any \textit{keywords} it considers too technical and unrelated to the main topics in $\AbstractAdv$ (the prompt can be found in Appendix \ref{appx:insert_keywords_prompt}).}
        \EndIf
        \vspace{1mm}
        \State $\AbstractAdvVersion{i} \gets \AbstractAdv$ 
    \EndFor

    \State \textbf{return} $\AbstractAdvVersion{i}$ that has highest $\SimilarityAbstract{\Title}{\AbstractAdvVersion{i}}{\ArchiveAdv{}}$
    \EndFunction
    \end{algorithmic}
\end{algorithm}

\subsubsection{\FindKeywords{} Subroutine}\label{appx:find_keywords}
In Algorithm \ref{alg:keywords}, we propose the \FindKeywords{} subroutine, which is a greedy search to find \textit{keywords} that when inserted into $\AbstractAdv$ raise the similarity to the $\ArchiveAdv{}$. To make this search efficient, we narrow down keywords by a heuristic that measures the increase in similarity upon appending the word to the current $\AbstractAdv$. This is just one simple instantiation of a possible attack strategy that uses the openly available SPECTER weights---more sophisticated attacks are possible, but we find that our simple heuristics are already extremely successful at breaking current systems. 
\begin{algorithm}[h!]
    \caption{\FindKeywords{} subroutine}\label{alg:keywords}
    \hspace*{\algorithmicindent} \textbf{Input:} \parbox[t]{.80\linewidth}{Paper title $\Title$, adversarial abstract from \IncludeRelatedWork{} $\AbstractAdv$, adversarial archive $\ArchiveAdv{}$, $\Keywords$ (number of keywords to return), and optionally \Call{Filter}{$\cdot$} (a function to filter out undesirable keywords)}\\
    \hspace*{\algorithmicindent} \textbf{Output:} \parbox[t]{.80\linewidth}{Up to $\Keywords$ keywords greedily selected to maximize the SPECTER similarity to $\AbstractAdv$ when the keywords are inserted into the abstract.}
    \vspace{0.1cm}
    \begin{algorithmic}[1]
    \Function{\FindKeywords{}}{$\Title, \AbstractAdv, \ArchiveAdv{}, \Keywords$}
    \State $\ArchiveWords \gets$ All words in titles and abstracts from $\ArchiveAdv{\Reviewer}$.
    \State $\ArchiveWords \gets$ \Call{Filter}{$\ArchiveWords$} \Comment{Optionally filter out certain words, e.g. numbers}
    \State \textit{keywords} $\gets$ {\tt []} \Comment{Keeps track of the keywords}
    \State \textit{keywordsSimilarity} $\gets$ $\SimilarityAbstract{\Title}{\AbstractAdv}{\ArchiveAdv{}}$ \Comment{Keeps track of estimated similarity to $\Reviewer$}
    \For{$i=0,\ldots,\Keywords-1$} \Comment{Iteratively select up to $\Keywords$  keywords}
        \For{each word $w_j$ in $\ArchiveWords$} \Comment{Simulate real modified abstracts with different $w_j$ added}
            \vspace{1mm}
            \State \parbox[t]{.80\linewidth}{$\AbstractAdvVersion{j} \gets$ adversarial abstract with words appended at the end ``\{$\AbstractAdv{}$\} \{$\ithMaxSimWord{0}$\} ... \{$\ithMaxSimWord{i-1}$\} \{$w_j$\}''.}
            \vspace{1mm}
        \EndFor
        \vspace{0.1cm}
        \State Let $\ithMaxSimWord{i}$ be a word $w_j$ such that the associated  $\SimilarityAbstract{\Title}{\AbstractAdvVersion{j}}{\ArchiveAdv{}}$ is the highest in $\ArchiveWords$.
        
        \If{$\underset{j}{\max} \ \SimilarityAbstract{\Title}{\AbstractAdvVersion{j}}{\ArchiveAdv{}} <$ \textit{keywordsSimilarity}} 
        \Comment{Stop if similarity does not increase}
            \State \textbf{break}
        \EndIf
        \State \textit{keywords}.append($\ithMaxSimWord{i}$)  \Comment{Add $\ithMaxSimWord{i}$ as a new keyword} 
        \State \textit{keywordsSimilarity} $\gets \underset{j}{\max} \ \SimilarityAbstract{\Title}{\AbstractAdvVersion{j}}{\ArchiveAdv{}}$  \Comment{Update estimated similarity with $\ithMaxSimWord{i}$ added}
    \EndFor
    \State \textbf{return} \textit{keywords}
    \EndFunction
    \end{algorithmic}
\end{algorithm}

\clearpage
\section{Attack Budgets} \label{sec:budgets}
We evaluate success rates under varying attack budgets by studying how the hyperparameters of \IncludeRelatedWork{} and \InsertKeywords{} operations ($\Regenerations$, $\Alternations$, $\Keywords$) affect automatic attack success rates of 50 samples with natural rankings of 101 from the NeurIPS 2023 dataset.

\begin{figure}[h]
    \centering
    \includegraphics[scale=0.37]{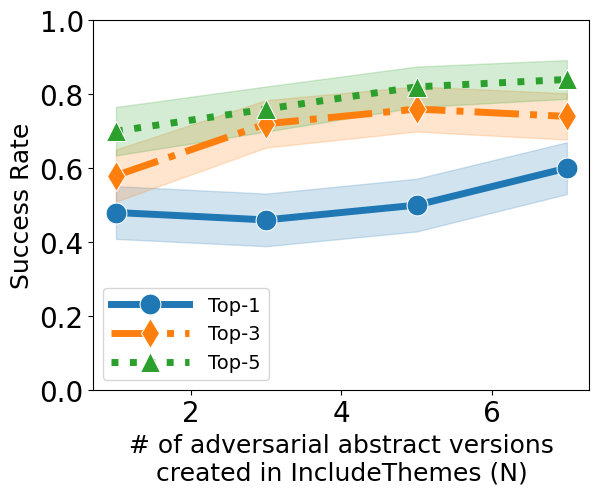}
    \caption{Attack success rates generally increase with $\Regenerations$ when the most similar attempt is kept out of $\Regenerations$ versions of $\AbstractAdv$ in \IncludeRelatedWork{}. The band is the standard error.}
    \label{fig:regen}
\end{figure}

First, Figure~\ref{fig:regen} shows attack success rates for varying $\Regenerations$—the number of \IncludeRelatedWork{} rounds— without using \InsertKeywords{} ($\Alternations=0$, $\Keywords=0$). Success rates generally increase with $\Regenerations$. This is expected because language model embeddings can be sensitive to paraphrasing, so taking the most-similar attempt amongst stochastic outputs helps improve attack success rates. However, there are signs of diminishing returns as $\Regenerations$ increases. 

\begin{figure}[h]
\centering
\begin{subfigure}[t]{.29\linewidth}
    \centering\includegraphics[trim={0 0 0 28},clip,width=1\linewidth]{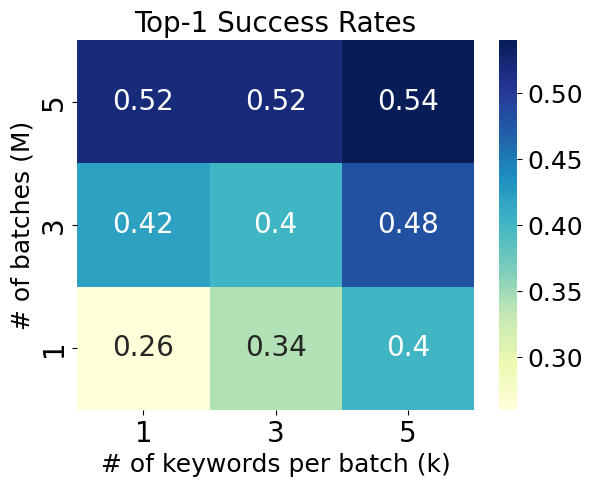}
    \caption{Top-1}
\end{subfigure}
\begin{subfigure}[t]{.29\linewidth}
    \centering\includegraphics[trim={0 0 0 28},clip,width=1\linewidth]{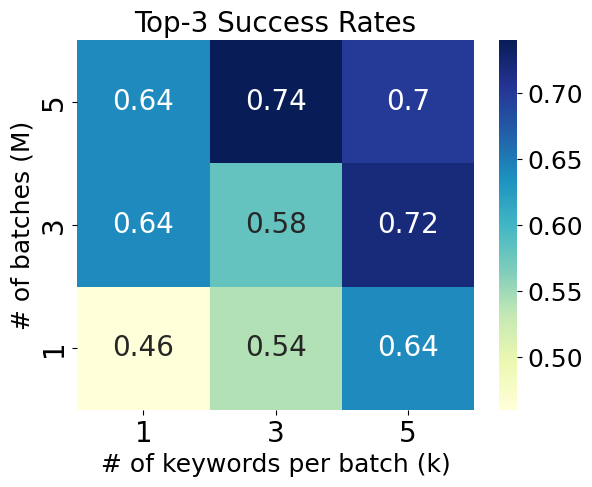}
    \caption{Top-3}
\end{subfigure}
\begin{subfigure}[t]{.29\linewidth}
    \centering\includegraphics[trim={0 0 0 28},clip,width=1\linewidth]{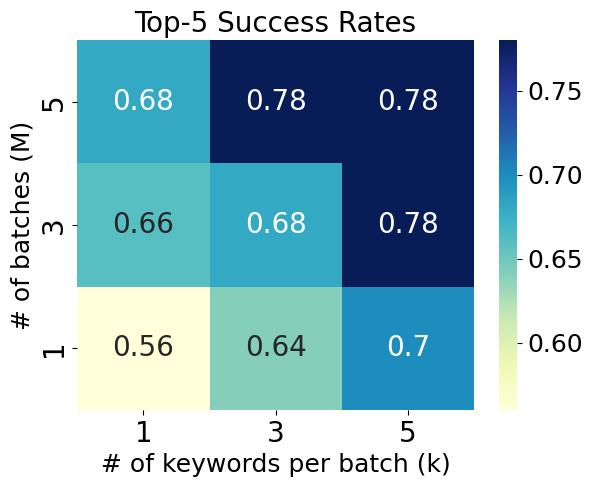}
    \caption{Top-5}
\end{subfigure}
\caption{Grids of Top-1, Top-3, and Top-5 success rates for different combinations of batches of keywords to insert in \InsertKeywords{} ($\Alternations$) and the number of keywords per batch returned by \FindKeywords{} ($\Keywords$).}
\label{fig:Sk-budgets}
\end{figure}

Second, in Figure~\ref{fig:Sk-budgets}, we report attack success rates for varying $\Alternations$ and $\Keywords$ in the \InsertKeywords{} operation and \FindKeywords{} subroutine, without using \IncludeRelatedWork{} ($\Regenerations=0$). The success rates increase with the total number of inserted keywords, $\Alternations \times \Keywords$, as shown by the color gradient from bottom left to top right in each subfigure. For a fixed keyword count, inserting smaller batches over more \InsertKeywords{} iterations outperforms larger batches over fewer iterations, as seen in the contrast between upper-left and bottom-right squares in those subfigures. 
This supports the intuition that gradual updates help greedy search adapt to changes in $\AbstractAdv{}$ from earlier batches.

\section{Similarity Scores and Rankings}\label{appx:natural_sim_and_rankings}

\begin{figure}[tb]
\centering
\begin{subfigure}[tb]{.45\linewidth}
    \centering\includegraphics[width=0.6\linewidth]{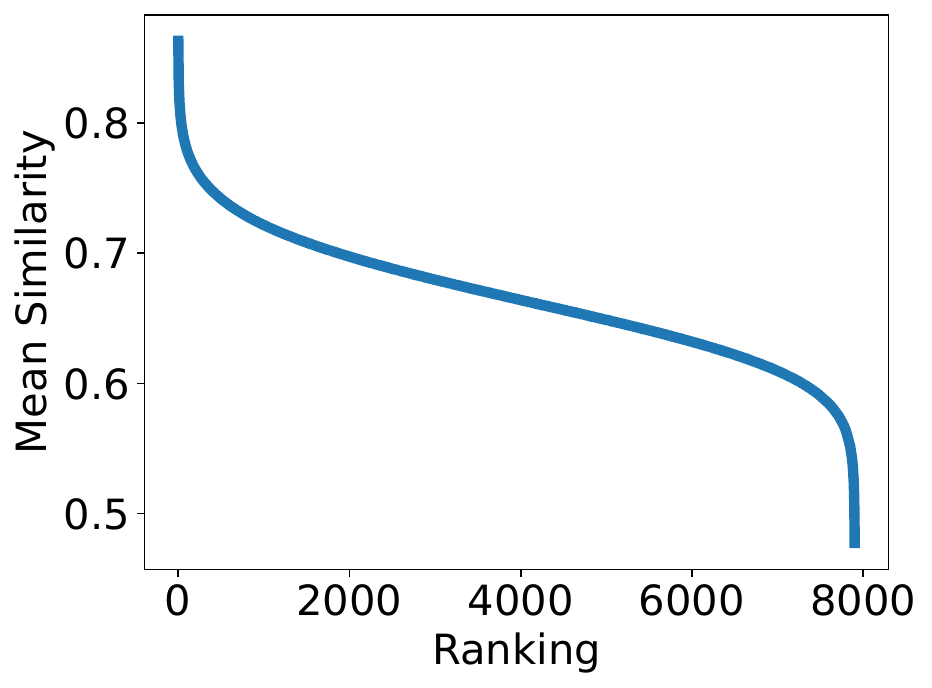}
    \caption{Mean similarity scores associated with each ranking from 1 to 7900. The shaded area represents standard error but is too small to be visible.} \label{subfig:all-rankings}
\end{subfigure} 
\hspace{0.03\linewidth}
\begin{subfigure}[tb]{.45\linewidth}
    \centering\includegraphics[width=0.6\linewidth]{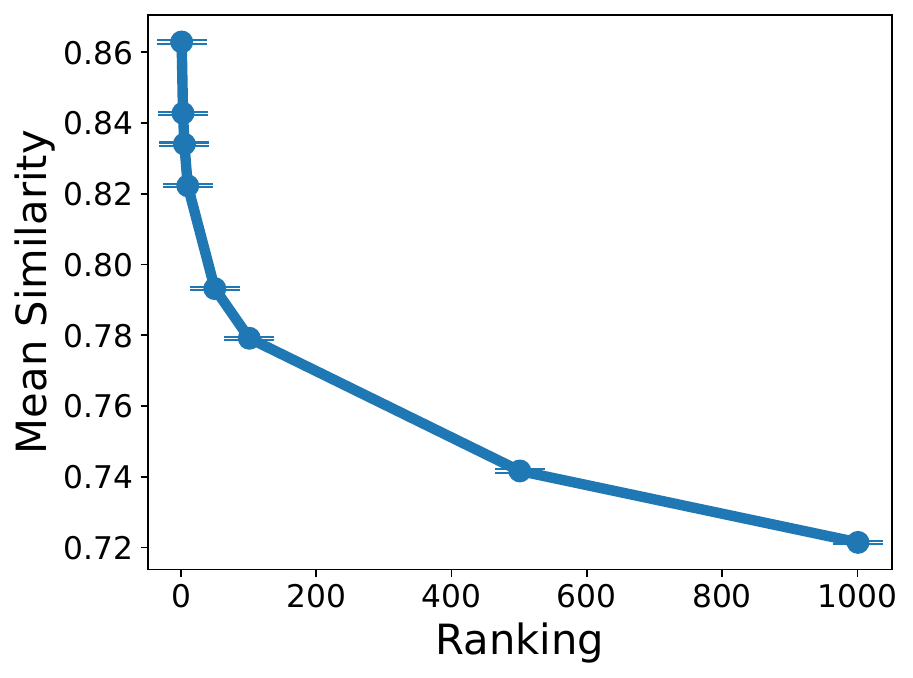}
    \caption{Zoomed in plot of the mean similarity scores, only for selected rankings between 1 and 1001. Error bars represent standard errors.} \label{subfig:1001-rankings}
\end{subfigure}
\caption{Mean similarity scores (before attack) associated with different rankings across all papers in the NeurIPS 2023 dataset. The curve exhibits a steep decline at the highest and lowest ranks, indicating that only a small fraction of reviewers has exceptionally high or low similarity to a given paper.}
\label{fig:similarities}
\end{figure}

In this work, we evaluate attack success using natural and manipulated rankings, as relative similarity rankings are more broadly applicable across conferences that may use different optimization programs for reviewer assignment (Section~\ref{sec:assignments}). To relate natural rankings to absolute similarity, Figure~\ref{fig:similarities} plots the mean similarity score (before attack) associated with each ranking across all papers in the NeurIPS 2023 dataset. For each paper, we rank all non-author reviewers by similarity and average the scores at each rank position over all papers.

Figure~\ref{subfig:all-rankings} shows that only a small fraction of reviewers have exceptionally high or low similarity scores to each paper, while mid-ranking reviewers exhibit a much more gradual decline in similarity scores. Figure~\ref{subfig:1001-rankings} zooms in on selected rankings between 1 and 1001 for closer inspection.

\section{Similarity Scores and Attack Results}\label{appx:success_stratified}

Some papers--such as those in niche fields--may have fewer high‐similarity reviewers, while others may have more. As a result, even colluding reviewer-paper pairs with identical ranking positions may have different values of similarity scores, leading to varying attack success rates. In this section, we explore how those natural similarity scores affect the likelihood of a successful attack.

For each paper, we calculate the similarities to all reviewers and then rank them. This allows us to compute the top and bottom quartiles with respect to the similarity scores of all reviewer-paper pairs with the same natural ranking. In particular, we focus on rankings 20, 101, 501, and 1001. We then revisit the main results from Section~\ref{sec:automatic_result}, applying this new stratification. For each ranking, we consider the colluding pairs identified in Section~\ref{sec:automatic_result} and categorize them into two groups: those whose natural similarity falls in the bottom quartile and those in the top quartile. In Table~\ref{tab:automatic_success_quartile}, success rates are generally lower for the bottom quartile group, though attacks can still achieve considerable success in these cases. We note that results are mixed when the natural ranking is 20, likely because attacks are highly successful for both groups.

\begin{table}[tb]
\caption{Attack success rates of groups of colluding pairs with natural similarity (before attack) in the top and bottom quartiles for their respective natural rankings. The baseline attack success rates from Section~\ref{sec:automatic_result} are also provided.}
\centering
\renewcommand{\arraystretch}{1.2}
\begin{tabulary}{\linewidth}{@{}CCCCC@{}}
 \toprule
 & & \multicolumn{3}{c}{Attack Success Rates ($\pm$ SE)} \\ \cmidrule(lr){3-5}
 \mbox{Natural} & \multirow{2}*{\mbox{Top-K}} & From & \multirow{2}*{Bottom quartile} & \multirow{2}*{Top quartile} \\ 
 \mbox{Ranking} &  & \mbox{Section \ref{sec:automatic_result}} & & \\\midrule

  20 & Top-1 & \mbox{\SI{90 \pm 3}{\percent}} & \SI{87 \pm 7}{\percent} & \SI{88 \pm 6}{\percent} \\
     & Top-3 & \mbox{\SI{96 \pm 2}{\percent}} & \SI{96 \pm 4}{\percent} & \SI{94 \pm 4}{\percent} \\
     & Top-5 & \mbox{\SI{98 \pm 1}{\percent}} & \SI{100 \pm 0}{\percent} & \SI{94 \pm 4}{\percent} \\ \midrule
     
 101 & Top-1 & \mbox{\SI{74 \pm 3}{\percent}} & \SI{56 \pm 6}{\percent} & \SI{86 \pm 4}{\percent} \\
     & Top-3 & \mbox{\SI{89 \pm 2}{\percent}} & \SI{77 \pm 5}{\percent} & \SI{97 \pm 2}{\percent} \\
     & Top-5 & \mbox{\SI{93 \pm 2}{\percent}} & \SI{82 \pm 5}{\percent} & \SI{99 \pm 1}{\percent} \\ \midrule

 501 & Top-1 & \mbox{\SI{60 \pm 5}{\percent}} & \SI{46 \pm 10}{\percent} & \SI{78 \pm 9}{\percent} \\
     & Top-3 & \mbox{\SI{76 \pm 4}{\percent}} & \SI{61 \pm 9}{\percent} & \SI{83 \pm 8}{\percent} \\
     & Top-5 & \mbox{\SI{83 \pm 4}{\percent}} & \SI{68 \pm 9}{\percent} & \SI{83 \pm 8}{\percent} \\ \midrule

 1001 & Top-1 & \mbox{\SI{48 \pm 5}{\percent}} & \SI{33 \pm 10}{\percent} & \SI{48 \pm 11}{\percent} \\
      & Top-3 & \mbox{\SI{63 \pm 5}{\percent}} & \SI{58 \pm 10}{\percent} & \SI{65 \pm 10}{\percent} \\
      & Top-5 & \mbox{\SI{67 \pm 5}{\percent}} & \SI{58 \pm 10}{\percent} & \SI{74 \pm 9}{\percent} \\

 \bottomrule
\end{tabulary}
\label{tab:automatic_success_quartile}
\end{table}

\newpage
\section{LLM Prompts}
\subsection{\IncludeRelatedWork{} Prompts} \label{appx:include_themes_prompt}
There are two different prompts for the \IncludeRelatedWork{} operation.

\subsubsection{Human-in-the-loop Mode} \label{appx:include_themes_manual_prompt}
For the ``human-in-the-loop'' mode, we use the following prompt to ask the LLM to generate abstracts that include themes from the colluding reviewer's archive (referred to as ``related previous works'' in the prompt). We manually edit the LLM-generated abstract to make sure the coherence and consistency constraints are satisfied. 

\vspace{2mm}
\fbox{
    \begin{minipage}{0.85\linewidth}
     \textbf{\IncludeRelatedWork{} Prompt (\HumanMode{} mode):} Please help me re-write my current academic abstract to add a short introduction under 40 words, relating my work to the concepts in previous works provided in a list. Do not include titles of the previous works in my abstract. I will provide you with a JSON dictionary with the following structure:

    \{
    ``title": my paper's title,
    ``abstract": my paper's abstract
    ``related previous works": [
        {"title": title1, "abstract": abstract1},
        {"title": title2, "abstract": abstract2},
        ....
    ]\} 
    \end{minipage}
}
\vspace{2mm}

\subsubsection{Automatic Mode} \label{appx:include_themes_auto_prompt}
In the ``automatic'' mode, we tuned the prompt more carefully to reduce incoherent or inconsistent abstract generations. First, we ask the LLM to add only \emph{one} additional sentence about the colluding reviewer's archive. In addition, we observe that the LLM tend to make scientifically false statements, such as falsely claiming that algorithms from the colluding reviewer's archive are also used in the paper with the manipulated abstract. Therefore, we ask the LLM to follow a format: explain that my work is \emph{inspired by} the themes in the previous work. For the writing style, we also instruct the LLM to use a matter-of-fact writing tone that is common for scientific publications. Lastly, we ask the LLM to remove personally identifiable information in the abstract, since most conferences anonymize authors during peer review.

\vspace{2mm}
\fbox{
    \begin{minipage}{0.85\linewidth}
     \textbf{\IncludeRelatedWork{} Prompt (\AutoMode{} mode):} In order for my paper to reach certain audiences, having the right topics in the abstract is very important. Edit the abstract to add one sentence to the introduction, explaining that my work is inspired by the themes in the previous works provided in a list.
    
    Here are a few requirements when writing the abstract:
    \begin{enumerate}
        \item Do not include titles of the previous works in my abstract.
        \item Use a matter-of-fact writing style common for scientific publications and avoid adjectives. Please especially avoid hyping up research with adjectives such as ``burgeoning'', ``transformative'', ``groundbreaking'', etc.
    \end{enumerate}

    Finally, please remove any personal identifiable information, such as GitHub links, from my abstract.
    
    I will provide you with a JSON dictionary with the following structure:
    \{
    ``title": my paper's title,
    ``abstract": my paper's abstract
    ``related previous works": [
        {``title": title1, ``abstract": abstract1},
        {``title": title2, ``abstract": abstract2},
        ....
        ]
    \} 

    Format your answer into JSON with the following schema:
    \{
    ``title": title string (should be the same as original),
    ``abstract": edited abstract, relating my work to the concepts in the 
                previous works provided. 
    \}
    \end{minipage}
}
\vspace{2mm}

For both human-in-the-loop and manual modes, we provide two examples to the LLM to guide its generation. The examples can be found on Zenodo \url{https://doi.org/10.5281/zenodo.15588237}.

\subsection{The \InsertKeywords{} Prompt} \label{appx:insert_keywords_prompt}
For the \InsertKeywords{} operation, we manually insert the keywords in the ``human-in-the-loop'' mode without the help of LLMs. In the ``automatic'' mode, we use the following prompt to ask the LLM to insert the keywords.

\vspace{2mm}
\fbox{
    \begin{minipage}{0.85\linewidth}
     \textbf{\InsertKeywords{} Prompt (\AutoMode{} mode):} In order for my paper to reach certain audiences, having the right keywords in the abstract is very important. I will provide you with a JSON dictionary with three keys: ``title'', ``abstract'' and ``keywords''. I want you to insert each keyword to the abstract based on its meanings commonly used in general English or meanings related to the technical details in the abstract. 

    Here are a few requirements when writing the abstract:
    \begin{enumerate}
        \item You must write a professional and scientifically rigorous abstract. Use a matter-of-fact tone.
        \item Use well-known facts in the scientific community when inserting keywords. Do not make changes to the parts related to this specific paper. 
        \item Some keywords may already exist in the abstract, but you must repeat the keyword somewhere else in the abstract. 
    \end{enumerate}
    
    Finally, some keywords are out of the scope of the abstract. You may reject them and provide a short 20-word explanation of why.

Format your answer into JSON with the following schema:
\{
    ``title": title string (should be the same as original),
    ``abstract": edited abstract string,
    ``left out keywords": {
        first rejected keyword: 20-word explanation of why the keyword 
                                is rejected.
        ...
    }
\}
    \end{minipage}
}
\vspace{2mm}

For this operation, we also provide two examples to the LLM to guide its generation. The examples can be found on Zenodo \url{https://doi.org/10.5281/zenodo.15588237}.

\subsection{Prompts used in Section \ref{sec:human_subject_experiment} (Human Subject Experiment)} \label{appx:hse_prompts}
As mentioned in Section \ref{sec:human_subject_experiment}, the prompts used for automatic abstract modification in the human subject experiment is tuned separately on the NeurIPS 2023 test data. Although in different words, the \IncludeRelatedWork{} prompt here has the same rules as the automatic mode prompt presented in Appendix \ref{appx:include_themes_auto_prompt} about the number of sentences to add, the \emph{inspired by} format to follow, and using a scientific writing style. Similarly, for the \InsertKeywords{} prompt, the rules here and in Appendix \ref{appx:insert_keywords_prompt} about how to insert the keywords, use professional writing styles, and not to add completely unrelated keywords are the same.

\vspace{2mm}
\fbox{
    \begin{minipage}{0.85\linewidth}
     \textbf{\IncludeRelatedWork{} Prompt (human subject experiment, \AutoMode{} mode):} Please help me edit my abstract's introduction to explain in one sentence that my work is inspired by the list of previous works in the provided JSON dictionary under the key "related previous works". Do not include the titles of previous works.

    Use the usual writing style of technical academic abstracts, avoid exaggerations and figurative language. Do not use flowery words or phrases such as ``prowess''. In addition, please remove any identifiable information (e.g. GitHub URLs) in my abstract by simply replacing them with [omitted for de-identification].

    I will provide you with a JSON dictionary with the following structure:
    \{
    "title": my paper's title,
    "abstract": my paper's abstract
    "related previous works": [
        {"title": title1, "abstract": abstract1},
        ....
    ]
    \}

    Format your answer into JSON with the following schema:
    \{
    "title": title string (should be the same as original),
    "abstract": edited abstract with short introduction, explaining my 
                work is inspired by the previous works provided
    \}
    \end{minipage}
}

\vspace{2mm}
\fbox{
    \begin{minipage}{0.85\linewidth}
     \textbf{\InsertKeywords{} Prompt (human subject experiment, \AutoMode{} mode):} Please help me edit my academic paper abstract to include a few provided keywords. Use the usual writing style of technical academic abstracts, avoid exaggerations and figurative language. Do not use flowery words or phrases such as "prowess". I will provide you with a JSON dictionary with three keys: ``title'', ``abstract'' and ``keywords''. 

    I want you to insert each keyword provided in the JSON to the abstract based on its meanings commonly used in general English or meanings related to the technical details in the abstract. Avoid inserting the words to the first or the last sentences of the abstract. In addition, please do not make changes to the title. 
    
    Some keywords are not commonly used in English and are not technically related to the main topics of the paper; please exclude them and provide a short 20-word explanation of why the keyword is unrelated to my abstract. However, you must insert a keyword that carries broad and general meanings; you cannot exclude it.

Format your answer into JSON with the following schema:
\{
    "title": title string (should be the same as original),
    "abstract": edited abstract string,
    "left out keywords": {
        first rejected keyword: 20-word explanation of why the keyword 
                                is rejected.
        ...
    }
\}

    If a keyword is not excluded as ``left out keywords", you must add it to the edited abstract. When inserting each keyword, you should use either technical or commonly used English meanings. Please add more instances of the keywords that are already present in the submission's abstract.
    \end{minipage}
}
\vspace{2mm}

\end{document}